\documentclass{article}

\usepackage{microtype}
\usepackage{graphicx}
\usepackage{subcaption}
\usepackage{booktabs}
\usepackage{multirow}
\usepackage{hyperref}

\usepackage{todonotes}
\usepackage{autonum}

\usepackage{algorithm}
\usepackage{algorithmic}

\AtBeginDocument{ %
\def\[#1\]{\begin{align}#1\end{align}}
}

 \usepackage[accepted]{icml2026}

\usepackage{amsthm}
\usepackage{amsmath}
\usepackage{amssymb}
\usepackage{mathtools}

\def\argmin{\operatornamewithlimits{arg\,min}}

\newcommand{\samplesize}{n}
\newcommand{\obsspace}{\mathcal{X}}
\newcommand{\randidx}{I}

\newcommand{\obsidx}{i}

\newcommand{\latvar}{z}

\newcommand{\ob}{X}
\newcommand{\obs}[1]{{\ob_{#1}}}
\newcommand{\prior}{\pi_{0}}
\newcommand{\globpar}{\theta}

\newcommand{\limglob}{\vartheta}

\newcommand{\innerprod}[2]{\left<#1,#2\right>}
\newcommand{\limlat}{\zeta}

\newcommand{\invtemp}{\beta}
\newcommand{\batchsize}{b}

\newcommand{\batchsizen}{\batchsize^{(\subseq)}}
\newcommand{\stepsize}{h}

\newcommand{\stepsizen}{\stepsize^{(\subseq)}}

\newcommand{\spascaln}{w^{(\subseq)}}

\newcommand{\timescalparn}{\alpha^{(\subseq)}}
\newcommand{\noise}{\xi}

\newcommand{\mleglob}{\hat{\globpar}}
\newcommand{\mleglobno}{\mleglob^{(n)}}
\newcommand{\mleglobn}{\mleglob^{(\subseq)}}
\newcommand{\conststep}{c_{\stepsize}}
\newcommand{\constbatch}{c_{\batchsize}}

\newcommand{\powerstep}{\mathfrak{h}}
\newcommand{\powerbatch}{\mathfrak{b}}
\newcommand{\powerit}{\mathfrak{t}}
\newcommand{\powerspa}{\mathfrak{w}}
\newcommand{\powertime}{\mathfrak{a}}

\newcommand{\globincrenoise}{\Delta_{\noise}^{(\subseq)}(\limglob)}
\newcommand{\globincreprior}{\Delta_{\prior}^{(\subseq)}(\limglob)}
\newcommand{\globincrell}{\Delta_{\ell}^{(\subseq)}(\limglob)}
\newcommand{\globincre}{\Delta^{(\subseq)}(\limglob)}
\newcommand{\newlat}{\tilde{\zeta}}

\newcommand{\op}{A^{(\subseq)}}
\newcommand{\gen}{A}
\newcommand{\expect}{\mathbb{E}}
\newcommand{\precon}{\Gamma}

\newcommand{\subseq}{n_m}

\newcommand{\defas}{:=}

\usepackage[capitalize,noabbrev]{cleveref}

\crefname{lemma}{Lemma}{Lemmas}
\crefname{corollary}{Corollary}{Corollaries}
\crefname{theorem}{Theorem}{Theorems}
\crefname{assumption}{Assumption}{Assumptions}
\crefname{proposition}{Proposition}{Propositions}
\crefname{algorithm}{Algorithm}{Algorithms}

\theoremstyle{plain}
\newtheorem{theorem}{Theorem}[section]
\newtheorem{proposition}[theorem]{Proposition}
\newtheorem{lemma}[theorem]{Lemma}
\newtheorem{corollary}[theorem]{Corollary}
\theoremstyle{definition}

\newtheorem{assumption}[theorem]{Assumption}
\theoremstyle{remark}

\icmltitlerunning{Large-scale UQ for LVMs Using Subsampling MCMC}
\newcommand{\new}[1]{#1}
\newcommand{\delete}[1]{}
\begin{document}

\twocolumn[
\icmltitle{Large-scale Uncertainty Quantification for Latent Variable Models \\ Using Subsampling Markov Chain Monte Carlo}

\begin{icmlauthorlist}
  \icmlauthor{Xiaoyu Wang}{MS}
  \icmlauthor{Jonathan Huggins}{MS,CDS}
\end{icmlauthorlist}

\icmlaffiliation{MS}{Department of Mathematics \& Statistics, Boston University}
\icmlaffiliation{CDS}{Faculty of Computing \& Data Sciences, Boston University}

\icmlcorrespondingauthor{Xiaoyu Wang}{shawnwxy@bu.edu}

\vskip 0.3in
]

\printAffiliationsAndNotice{}  %

\begin{abstract}
Stochastic gradient Langevin dynamics combined with Gibbs updates (SGLD--Gibbs)
provides a highly scalable approach to approximate Bayesian inference in latent variable models. 
However, it remains unclear how to tune the algorithm's hyperparameters in a principled manner to ensure the uncertainty estimates are statistically meaningful.
In this work, we address this gap in tuning guidance by developing a statistical scaling limit theory for SGLD--Gibbs. We derive a joint asymptotic limit for the global parameters and latent
variables under appropriate space-time rescaling. 
We show that global parameters converge to a diffusion-type limit, while
each latent variable converges to a jump process, reflecting the use of intermittent Gibbs updates.
This joint jump-diffusion structure reveals how latent-variable randomness
contributes to the stationary distribution of the global parameters.
We leverage our results to propose explicit guidance on hyperparameter tuning for SGLD--Gibbs that ensures meaningful uncertainty quantification.
Numerical experiments show that SGLD--Gibbs with our tuning guidance leads to better parameter estimates, uncertainty quantification, and predictive performance than stochastic variational inference. 
\end{abstract}

\section{Introduction}

Stochastic gradient methods such as stochastic gradient descent (SGD) and
stochastic gradient Langevin dynamics (SGLD) have become central tools for
large-scale optimization and approximate Bayesian inference
\citep{Nemirovski2009,NIPS2011_40008b9a,bottou2016,conf/icml/WellingT11,sgmcmc2021}.
For approximate sampling, latent variable models (LVMs) are one of the most frequently cited applications of SG(L)D.
Examples include Gaussian mixture models, mixed-membership stochastic block
models \citep{MMSB}, latent Dirichlet allocation \citep{LDA}, Bayesian matrix
factorization \citep{BayMatFac}, mixed effects models \citep{Peter2023OMA}, and
discrete choice models \citep{Loaiza-Maya02102023,LOAIZAMAYA2024105741}.
In these applications, an \emph{SGLD--Gibbs} scheme is often used,
where SGLD update steps are constructed using one or more
conditional draws of the latent variables, enabling approximate posterior sampling with per-iteration costs
that scale favorably with data size.
However, there is little rigorous guidance on tuning the algorithmic hyperparameters
of SGLD--Gibbs such
as the step size, minibatch size, and inverse temperature.
Moreover, how to obtain meaningful uncertainty quantification using SGLD--Gibbs remains unclear.
A substantial body of work
\citep{Walk1977AnIP,pflug1986,kushner2003stochastic,negrea2022statistical,wang2025quantitativeerrorboundsscaling}
has employed scaling-limit analyses to study standard SG(L)D. 
These approaches relate SG(L)D sample paths to continuous-time stochastic
processes and yield characterizations of optimization accuracy, asymptotic
behavior, and numerical efficiency.
Such analyses have proven particularly useful for understanding hyperparameter
tuning and uncertainty quantification \citep{mandt2017stochastic,negrea2022statistical}.
However, existing scaling-limit results do not directly apply to latent variable
models.

In this work, we address this gap by jointly analyzing the global parameters and latent variables under appropriate space-time rescaling, which provides a unified asymptotic characterization of SGLD--Gibbs dynamics. 
We show that the global parameters converge to a diffusion-type limit, while
each latent variable converges to an independent jump process.
We further demonstrate that the interaction between the global diffusion and
latent-variable jumps fundamentally alters the noise structure of the
global parameters.
In particular, the latent variables contribute an
additional source of variability determined by the number of Gibbs samples used
per iteration. %
We use our results to derive concrete guidance for uncertainty quantification and
hyperparameter tuning.
Empirically, we find that SGLD--Gibbs yields improved accuracy and more reliable
uncertainty quantification compared to stochastic variational inference
in applications to mixture modeling and topic modeling. 

\subsection{Related Work and Alternative Approaches}

Given their widespread use, SG(L)D methods have been studied from many
perspectives, including finite-sample error bounds, convergence rates, and
stationary distributions
\citep[e.g.,][]{Mcleish1976,ruppert1988,doi:10.1137/0330046,kushner2003stochastic,
negrea2022statistical,rakhlin2011making,10.1214/19-AOS1850,pmlr-v125-mou20a,
cheng2020stochastic,srikant2024rates,pmlr-v99-anastasiou19a,ge2015escaping,jin2017escape}.
Most relevant to our work is scaling-limit theory for stochastic approximation algorithms, which shows that, under appropriate rescaling, SGD and
SGLD trajectories converge to Ornstein--Uhlenbeck diffusions \citep{kushner1981,kushner1993,kushner2003stochastic,negrea2022statistical}.
Further results characterize mixing times, stationary covariances, and the behavior of
averaged iterates \citep{mandt2017stochastic, negrea2022statistical,collins2024nips,qian2024sureconvergenceratesconcentration,kushner1993}.

Several recent works extend this theory to improve uncertainty quantification for
stochastic gradient algorithms.
\citet{wang2025quantitativeerrorboundsscaling} develop non-asymptotic functional
error bounds for diffusion approximations to scaling limits.
\citet{wang2026accurate} develop discrete-time proxy theories for stochastic
gradient algorithms, clarifying when diffusion-based uncertainty quantification
remains valid in large-batch or non-asymptotic regimes.
A separate line of work
studies scaling limits of SGD in high-dimensional regimes where the parameter
dimension $d \to \infty$, yielding mean-field or dynamical equations
for low-dimensional summary statistics \citep{nips2022arous,collinswoodfin2023hittinghighdimensionalnotesode,Mignacco_2021}.

Variational Bayesian methods, including mean-field variational Bayes, 
online variational Bayes, stochastic variational inference (SVI), 
and related variational approximations, can also provide scalable for latent variable model inference  \citep{JMLR:v14:hoffman13a,NIPS2010_OVB,ADVI2017}.
Nonetheless, their ability to quantify posterior uncertainty can be fundamentally
limited \citep{gelman2013bayesian,margossian2025variational,JASA2018giordano}.
For example, \citet{margossian2025variational} show that when the true posterior distribution exhibits dependence
structure, variational approximations based on factorization cannot, in general,
correctly estimate posterior uncertainty.
Depending on the divergence being minimized, uncertainty estimates produced by
variational Bayes are often poorly calibrated, even under correct model
specification.

\section{Preliminaries and Problem Setup}
\label{sec:preliminary}

This section introduces the class of latent variable models considered in this work, describes the SGLD--Gibbs algorithm, and reviews  preliminary results concerning scaling limits for stochastic gradient methods.

\subsection{Latent Variable Models}
We consider a general class of latent variable models in which each observation is associated with an unobserved latent variable.
Let $\{(X_i, z_i)\}_{i=1}^n$ denote independent pairs of observed data $X_i \in \mathcal X$ and latent variables $z_i \in \mathcal Z$.
The joint distribution is parameterized by a global parameter $\theta \in \Theta \subset \mathbb R^d$ and admits the factorization
\[
p(X_i, z_i \mid \theta) = p(z_i \mid \theta)\, p(X_i \mid z_i, \theta),
\]
with prior distribution $\pi_0(\theta)$ on $\theta$.
The marginal likelihood of the observations is given by
$
p(X_i \mid \theta) = \int p(X_i, z_i \mid \theta)\, dz_i,
$
and the corresponding log-likelihood is
$
\ell(\theta; X_i) := \log p(X_i \mid \theta).
$ 
This formulation encompasses a wide range of commonly used models, including mixture models,
mixed-membership stochastic block models, topic models, and Bayesian matrix factorization.
See \citet{pml2Book} for a systematic discussion of learning and approximate Bayesian inference in such models. 

\subsection{SGLD with Gibbs Updates}
\label{subsec:SGLD--Gibbs}
We study stochastic gradient Langevin dynamics combined with Gibbs updates for latent variables.
Let $b \in \{1,\dots,n\}$ denote the minibatch size.
At iteration \(k\), a minibatch of indices
\(I_k=\{I_k(1),\dots,I_k(b)\}\) is sampled uniformly from
\(\{1,\dots,n\}\) with replacement. \new{This convention is mainly for
theoretical convenience. Similar scaling-limit results for sampling
without replacement were established in \citet{negrea2022statistical}, and we
expect the conclusions here to remain the same under sampling without
replacement, up to constant-level deviations when \(b\) is of order \(n\).}
Given the current global parameter $\theta_k$, the algorithm proceeds in two steps:

\paragraph{(i) Gibbs updates of latent variables.}
For each $i\in I_k$, the latent variable is resampled from its conditional posterior,
\[
z_{i,k+1} \sim p(z_i \mid X_i, \theta_k),
\]
while latent variables not in the minibatch remain unchanged.

\paragraph{(ii) SGLD update of global parameters.}
Using the stochastic gradient estimator with refreshed latent variables given by
\[
\begin{split}
G_k(\theta)
&:=
\frac{1}{n}\nabla \log \pi_0(\theta) \\
&\phantom{:=~} + 
\frac{1}{b}\sum_{i\in I_k}
\nabla_\theta \log p\!\left(X_i, z_{i,k+1}\mid \theta\right),  \label{eq:sgld-gibbs-gradient}
\end{split}
\]
the global parameter is updated via
\[
\label{sgld_updates}
\theta_{k+1}
=
\theta_k
+
\frac{h}{2}\,\Gamma\, G_k(\theta_k)
+
\sqrt{\frac{h}{\beta}}\,\Gamma^{1/2}\xi_k,
\]
where $h>0$ is the step size, $\beta\in(0,\infty]$ is the inverse temperature, $\Gamma \in \mathbb R^{d\times d}$ is a positive definite preconditioning matrix, and $\xi_k\sim\mathcal N(0,I_d)$.
The full procedure is summarized in \cref{alg:sgld_gibbs}.

\begin{algorithm}[t] 
\caption{SGLD--Gibbs for Latent Variable Models}
\label{alg:sgld_gibbs}
\begin{algorithmic}[1]
\STATE \textbf{Input:} step size $h$, batch size $b$, inverse temperature $\beta$, preconditioner $\Gamma$, initial values $(\theta_0,\{z_{i,0}\}_{i=1}^n)$
\FOR{$k=0,1,2,\ldots$}
  \STATE Sample minibatch $I_k\subset\{1,\ldots,n\}$ with $|I_k|=b$
  \FOR{each $i\in I_k$}
    \STATE Sample $z_{i,k+1}\sim p(z_i\mid X_i,\theta_k)$
  \ENDFOR
  \STATE Update $\theta_{k+1}$ using the SGLD step with updated $\{z_{i,k+1}\}_{i\in I_k}$
\ENDFOR
\end{algorithmic}
\end{algorithm}

\subsection{Scaling Limits for Stochastic Gradient Methods}
\label{subsec:negreascalinglimit}

In a general setup that does not involve latent variables, assume observations are i.i.d.\ from an unknown distribution $P_{\star}$. 
The model is parameterized by a global parameter $\theta \in \Theta \subset \mathbb R^d$
and admits a likelihood of the form $p(X_i \mid \theta)$ for each observation $X_i$,
with prior distribution $\pi_0(\theta)$ on $\theta$.
The optimal parameter is given 
by $\theta_{\star} \defas \argmin_{\theta} \mathbb{E}\left[ \ell(X, \theta)\right]$, where $X \sim P_{\star}$. 

Recall that $I_k \subset \{1,\dots,n\}$ denotes the $k$th minibatch.
SGLD uses the one-step update given in  \cref{sgld_updates}, where the stochastic gradient estimator is now
\[
G_k(\theta)
:=
\frac{1}{n}\nabla \log \pi_0(\theta)
+
\frac{1}{b}\sum_{i\in I_k}
\nabla_\theta \log p\!\left(X_i\mid \theta\right).
\]

Scaling limit theory relates discrete-time stochastic gradient algorithms to continuous-time stochastic processes under appropriate space-time rescaling. Let $\theta^{(n)}_k \in \mathbb{R}^d$ denote the global parameter at iteration $k$ and $\hat\theta^{(n)}$ denote a critical point satisfying the first-order condition $\textstyle\sum_{i=1}^n \nabla \ell(\hat{\theta}^{(n)}; X_i) = 0$.
Define the rescaled, continuous-time process
\[
\label{eq:scaling_process}
\vartheta^{(n)}_t
=
n^{\mathfrak{w}}
\left(
\theta^{(n)}_{\lfloor n^{\mathfrak{a}} t \rfloor}
-
\hat{\theta}^{(n)}
\right),
\]
where $\mathfrak{w} > 0$ and $\mathfrak{a} > 0$ denote the spatial and temporal scaling exponents.

Then, under an appropriate scaling regime, this process converges in distribution
to an Ornstein--Uhlenbeck process whose drift and diffusion coefficients depend on the preconditioner $\precon$ and the first- and second-order Fisher information matrices
\[
I_\star
:=
\mathbb{E}
\left[
[\nabla_\theta \ell(\theta^\star; X)]^{\otimes 2}
\right],
J_\star
:=
-
\mathbb{E}
\left[
\nabla_\theta^{\otimes 2}
\ell(\theta^\star; X)
\right],
\]
\new{where \(a^{\otimes 2}:=a\otimes a\) denotes the outer product, and \(\nabla^{\otimes 2}\) denotes the Hessian operator.}

Here $I_\star$ quantifies the variability of the log-likelihood gradient at $\theta^\star$,
while $J_\star$ captures the local second-order behavior of the log-likelihood around $\theta^\star$.

\begin{theorem}[\citet{negrea2022statistical}, Theorem~1]
\label{thm:negrea_ou}
Consider the SGLD algorithm with step size
$h^{(n)} = c_h n^{-\mathfrak{h}}$,
batch size $b^{(n)} = \lfloor c_b n^{\mathfrak{b}} \rfloor$,
and inverse temperature
$\beta^{(n)} = c_\beta n^{\mathfrak{t}}$.
Let $\mathfrak{a}= \mathfrak{h}$ and $\mathfrak{w} = \min \{\mathfrak{b}+\mathfrak{h},\mathfrak{t}\}/2$.
Then, under mild regularity conditions, as $n \to \infty$,
\[\vartheta^{(n)}_t \Rightarrow \vartheta_t
\]
in the Skorohod topology in probability, where $\Rightarrow$ denotes weak convergence and
$\vartheta_t$ is an Ornstein--Uhlenbeck process solving
\[\label{eq:negrea_ou}
d\vartheta_t
=
-
\frac{1}{2} B \vartheta_t\, dt
+
\sqrt{A}\, dW_t,
\]
with
\[
\begin{aligned}
B
&=
c_h \Gamma J_\star \mathbf{1}\{ \mathfrak{a} = \mathfrak{h} \},\\
A
&=
\frac{c_h}{c_\beta} \Gamma \mathbf{1}\{ \mathfrak{h}+\mathfrak{b}\leq \mathfrak{t} \}
+
\frac{c_h^2}{4c_b}
\Gamma I_\star \Gamma^\top
\mathbf{1}\{ \mathfrak{t} \leq \mathfrak{b}+\mathfrak{h} \}.
\end{aligned}
\]
\end{theorem}

As discussed by \citet{negrea2022statistical}, one implication of this result is that characterization of 
full-path limiting dynamics relates the mixing behavior of SGLD to the spectral properties
of the limiting Ornstein--Uhlenbeck process.
Heuristically, the asymptotic mixing time then scales inversely to the smallest
eigenvalue of the corresponding drift matrix $B$.
This motivates choosing the preconditioner $\precon$ to approximate $J_\star^{-1}$, which
optimizes mixing speed in the asymptotic regime.

\subsection{Uncertainty Quantification}

\cref{thm:negrea_ou} shows how the scaling of the step size, minibatch size, and inverse temperature determines which noise sources remain active in the limiting diffusion, which determines the form of the stationary covariance, as summarized by the following standard result.

\begin{proposition}[\citet{negrea2022statistical}, Corollary~1]
\label{cor:stationary_covariance}
For the Ornstein--Uhlenbeck process $(\vartheta_t)_{t \ge 0}$ defined by
\eqref{eq:negrea_ou} with known $\vartheta_0$, 
the law of $\vartheta_t$ is Gaussian with mean
$e^{-tB/2} \vartheta_0$
and covariance
\[ \textstyle
\Sigma_t
=
\int_0^t
e^{-sB/2}
A
e^{-sB^\top/2}
\, ds.
\]
If a stationary distribution exists, then $\vartheta_t$ admits
$\mathcal{N}(0, \Sigma_\infty)$ as its stationary law, where
$\Sigma_\infty$ solves
\[  \textstyle
\frac{1}{2} B \Sigma_\infty
+
\frac{1}{2} \Sigma_\infty B^\top
=
A.
\]
\end{proposition}

Thus, Proposition~\ref{cor:stationary_covariance} enables the targeting of a desired stationary covariance  to ensure meaningful uncertainty quantification. 
In the Bayesian setting, the Bernstein-von Mises theorem states that the posterior is approximately $\mathcal{N}(\hat{\theta}^{(n)}, J_{\star}^{-1}/N)$ \citep{kleijn2012bernstein}.
Thus, one possible goal when using SGLD for uncertainty quantification is to obtain samples with a distribution that is approximately equal to $\mathcal{N}(\hat{\theta}^{(n)}, J_{\star}^{-1}/N)$.
However, the sampling distribution of $\hat{\theta}^{(n)}$ is asymptotically normal with mean $\theta_{\star}$ and covariance equal to $J_{\star}^{-1} I_{\star} J_{\star}^{-1}/N$ \citep{white1982maximum}.
The matrix $S_\star := J_{\star}^{-1} I_{\star} J_{\star}^{-1}$ is known as the ``sandwich'' covariance matrix, and it suggests that for proper uncertainty quantification
we want the stationary SGLD distribution to be approximately $\mathcal{N}(\hat{\theta}^{(n)}, J_{\star}^{-1} I_{\star} J_{\star}^{-1}/N )$.
When the model is correctly specified, $I_{\star} = J_{\star}$, so the sandwich covariance is equal to $J_{\star}^{-1}$ and the Bayesian
posterior provides correct uncertainty quantification.
In the misspecified setting, tuning SGLD to have stationary covariance $S_\star/N$ will correctly capture the sampling uncertainty. 
Alternatively, the model-based and sampling-based uncertainties can be combined, as in the bagged posterior \citep{huggins2024reproducible}.
Based on these statistical considerations, \citet{negrea2022statistical} propose to target either the Bernstein--von Mises uncertainty or the bagged posterior uncertainty, with the sampling uncertainty arising as a special case of the latter.
The recommended tunings derived from \cref{thm:negrea_ou} and Proposition~\ref{cor:stationary_covariance} are summarized in \cref{tab:tuning_recommendations}.

\begin{table}[t]
\centering
\begin{tabular}{lccccc}
\toprule
Target 
& Asymp. cov.
& $\Gamma$
& $\beta$
& $h$ \\
\midrule
BvM
& $J_\star^{-1}$
& $I_\star^{-1}$
& $\frac{n}{1-w_1}$
& $\frac{4w_1 b}{n}$.
\\

Bagged post.
& $w_2 J_\star^{-1} + w_1 S_\star$
& $J_\star^{-1}$
& $\frac{n}{w_2}$
& $\frac{4 w_1 b}{n}$ \\
\bottomrule
\end{tabular}
\caption{Recommended tuning parameter combinations 
and the corresponding asymptotic covariances of SGLD in \citet{negrea2022statistical}.
\textbf{BvM:} targeting the asymptotic covariance of the 
posterior based on the Bernstein--von Mises. 
\textbf{Bagged post.:} targeting the asymptotic covariance of the 
(generalized) bagged posterior \citep{huggins2024reproducible}. 
If targeting the sandwich covariance, use the bagged posterior tuning with $w_1 = 1$ and $w_2 = 0$. 
}
\label{tab:tuning_recommendations}
\end{table}

\section{Main Results}
\label{sec:main_results}

In this section we present our main theoretical results.
We first define the scaling-limit objects for both the global parameter and latent variables,
together with the associated \emph{latent-involved Fisher information matrix}.
We then state our main joint scaling-limit theorem for SGLD--Gibbs,
followed by corollaries on uncertainty quantification and latent-variable mixing.
All technical assumptions and proofs are deferred to the supplementary materials.

\subsection{Scaling-limit objects and information matrices}

We take the same definition of rescaled global-parameter process $\vartheta^{(n)}_t$ as in \eqref{eq:scaling_process} and define latent-variable process by
\[
\zeta^{(n)}_{i,t}
=
z^{(n)}_{i,\lfloor n^{\mathfrak{a}} t \rfloor},
\]
where $z^{(n)}_{i,k}$ denote the latent variable associated with observation $i$ at $k$th iteration.

Distinguished from $I_\star$ defined in \cref{subsec:negreascalinglimit}, we define \emph{latent-involved Fisher information matrix}
\[
\widetilde I_\star
:=
\mathbb{E}_{X,Z \mid \theta^\star}
\left[
\nabla_\theta \log p(X, Z \mid \theta^\star)^{\otimes 2}
\right]
=
I_\star
+
M_\star,
\]
where
\[
M_\star
:=
\mathbb{E}_{X\mid\theta^\star}
\Big[
\mathrm{Var}_{Z\mid X,\theta^\star}
\big(
\nabla_\theta \log p(X,Z\mid\theta^\star)
\big)
\Big] \succeq 0
\]
is the ``Jensen gap'' that quantifies the additional \emph{algorithm-induced uncertainty} due to estimating the marginal likelihood with only a single Gibbs sample. 
It follows from the definition of $M_\star$ that the gap will be 
smaller when, on average, there is less uncertainty about $Z$ given $X$. 

\subsection{Joint scaling limit for global and latent parameters}

We now show that, under an appropriate asymptotic scaling, the rescaled global parameter
and a fixed latent variable (e.g., $z_1$) converge jointly in distribution.
The limiting process has independent diffusion and jump components.
Specifically, the global parameter process converges to an Ornstein--Uhlenbeck process whose drift
and diffusion structure is analogous to that in \cref{thm:negrea_ou}, except that the diffusion
matrix is modified to explicitly incorporate algorithm-induced uncertainty arising in
SGLD--Gibbs.
Meanwhile, the latent-variable process converges to a Poisson-driven Gibbs jump process,
with jumps drawn from the true conditional posterior of the latent variable.
\begin{theorem}[Joint scaling limit of SGLD--Gibbs]
\label{thm:joint_scaling_limit}
Consider the SGLD--Gibbs algorithm with the same polynomial scaling of tuning parameters as in \cref{thm:negrea_ou}. 
Let $\mathfrak{a}= \mathfrak{h}$ and $\mathfrak{w} = \min \{\mathfrak{b}+\mathfrak{h},\mathfrak{t}\}/2$.
Assume the regularity conditions stated in \cref{app:assumptions}. 
Then, as $n \to \infty$,
\[
\bigl(\vartheta^{(n)}_t, \zeta^{(n)}_{1,t}\bigr)
\Rightarrow
\bigl(\vartheta_t, \zeta_{1,t}\bigr),
\]
in the Skorokhod topology in probability, where the limiting processes
$\vartheta_t$ and $\zeta_{1,t}$ are independent and defined as follows:
\begin{enumerate}
\item The global-parameter limit $\vartheta_t$ is an Ornstein--Uhlenbeck process solving
\[
d\vartheta_t
=
-
\frac{1}{2} B \vartheta_t\, dt
+
\sqrt{A}\, dW_t,
\]
with
\[
\begin{aligned}
B
&=
c_h \Gamma J_\star \mathbf{1}\{ \mathfrak{a} = \mathfrak{h} \},\\
A
&=
\frac{c_h}{c_\beta} \Gamma \mathbf{1}\{ \mathfrak{h}+\mathfrak{b}\leq \mathfrak{t} \}
+
\frac{c_h^2}{4c_b}
\Gamma \tilde{I_\star} \Gamma^\top
\mathbf{1}\{ \mathfrak{t} \leq \mathfrak{b}+\mathfrak{h} \}.
\end{aligned}
\]

\item When $ \mathfrak{h} +\mathfrak{b} \leq 1$, the latent-variable limit $\zeta_{1,t}$ is a pure-jump Markov process with generator
\[
\begin{aligned}
(\mathcal{L} f)(z)
&=
\lambda
\int
\bigl\{ f(z') - f(z) \bigr\}
\, p(z' \mid X_1, \theta^\star)\, dz',
\end{aligned}
\]
where $\lambda := c_b \mathbf{1}\{ \mathfrak{h} +\mathfrak{b} =1 \}.$
\end{enumerate}
\end{theorem}

For global parameters, the joint scaling limit in \cref{thm:joint_scaling_limit} yields the same scaling regimes in $(\mathfrak h,\mathfrak b,\mathfrak t)$ as 
\cref{thm:negrea_ou}.
The key difference in latent variable models is that the diffusion term involves $\widetilde I_\star$, which captures additional uncertainty arising from the latent variables and Gibbs sampling. This latent-induced contribution persists in the scaling limit and affects the stationary covariance of the global parameters, beyond what is implied by the marginal likelihood alone.

The behavior of the latent-variable dynamics depends critically on the relation
between the scaling exponents $\mathfrak h$ and $\mathfrak b$.
If $\mathfrak h + \mathfrak b = 1$, each latent variable evolves as a pure-jump
Markov process with a nondegenerate limiting intensity, yielding a meaningful
joint jump-diffusion limit as characterized in \cref{thm:joint_scaling_limit}.
If $\mathfrak h + \mathfrak b < 1$, \delete{latent variables are effectively frozen on the
macroscopic time scale, leading to persistent bias in the global-parameter dynamics.}\new{ the latent-variable dynamics become degenerate on the macroscopic timescale; this is because $z_1$ is refreshed too infrequently relative to the evolution of $\theta$, so the latent-variable process effectively freezes and loses any meaningful limiting behavior. }
If $\mathfrak h + \mathfrak b > 1$, latent-variable updates occur on a much faster
time scale than the global parameters.
In this scenario, the latent-variable generator diverges in the limit, and thus the
sequence of latent-variable processes does not admit a well-defined limit.

\begin{corollary}[Latent-variable limit: stationarity and mixing]
\label{cor:latent_mixing}
Assume the regime of Theorem~\ref{thm:joint_scaling_limit} with $\mathfrak h + \mathfrak b = 1$,
in which the latent-variable limit is a pure-jump Markov process with rate $\lambda = c_b$.
Let $\mu_t$ denote the law of $\zeta_{1,t}$ with initial law $\mu_0$.
Define the $\varepsilon$-mixing time in total variation by
\[
t_{\mathrm{mix}}(\varepsilon)
\defas
\inf\bigl\{ t \ge 0 : d_{\mathrm{TV}}(\mu_t,\pi) \le \varepsilon \bigr\}.
\]
Then $\pi(\cdot)=p(\,\cdot\mid X_1,\theta^\star)$ is the unique stationary distribution and,
for all $t\ge0$,
\[
\mu_t
=
e^{-\lambda t}\mu_0
+
(1-e^{-\lambda t})\pi .
\]
Consequently,
$
d_{\mathrm{TV}}(\mu_t,\pi)
=
e^{-\lambda t} d_{\mathrm{TV}}(\mu_0,\pi)
$,
and hence 
\[
t_{\mathrm{mix}}(\varepsilon)
=
\frac{1}{\lambda}
\log\!\left(\frac{1}{\varepsilon}\right).
\]
\end{corollary}
This result implies that $\zeta_{1,t}$ evolves as a refresh process that
periodically resamples from $p(z_1\mid X_1,\theta^\star)$.
Since each latent-variable update acts as an instantaneous
refresh in the scaling limit, the effective mixing time is of order $1/\lambda$.
\new{For the global parameter, the relevant mixing heuristic is inherited from \cref{{thm:negrea_ou}} via the drift matrix $B$ of the limiting Ornstein--Uhlenbeck process, as in \citet{negrea2022statistical} }

\begin{table*}[!t]
\centering
\begin{tabular}{lccccc}
\toprule
Target 
& Asymp.\ cov.
& $\Gamma$
& $\beta$
& $h$ \\
\midrule
Bernstein--von Mises
& $J_\star^{-1}$
& $\bigl(\widetilde I_\star^{(S)}\bigr)^{-1}$
& $n / (1-w_1)$
& $4 w_1 b / n$ \\

Bagged post.\ \{+ algorithmic variability\}
& $w_1 S_\star + w_2 J_\star^{-1} + \dfrac{w_1}{S}\, J_\star^{-1} M_\star J_\star^{-1}$
& $J_\star^{-1}$
& $n / w_2$
& $4 w_1 b / n$ \\
\bottomrule
\end{tabular}
\caption{Recommended tuning parameter combinations and the corresponding asymptotic
covariances of SGLD--Gibbs.
See the caption of \cref{tab:tuning_recommendations} for further explanation.
}
\label{tab:gibbs_sgld_tuning_recommendations}
\end{table*}

\subsection{Scaling limit under generalized Gibbs approximation}

We now consider a natural extension of the SGLD–Gibbs update that is commonly used in practice, in which the single latent-variable draw in the stochastic gradient is replaced by an average over multiple conditional samples. 
Letting $S$ denote the number of Gibbs samples per iteration, for each $i \in I_k$ now draw
$z_{i,k+1}^{(1)},\dots,z_{i,k+1}^{(S)} \stackrel{\text{i.i.d.}}{\sim}
p(z_i \mid X_i,\theta_k)$.
We then define the averaged stochastic gradient
\begin{align}
G_k^{(S)}(\theta)
&\defas
\frac{1}{n}\nabla \log \pi_0(\theta) \\
&\phantom{=~} +
\frac{1}{bS}\sum_{i\in I_k}
\sum_{s=1}^S
\nabla_\theta \log p\!\left(X_i, z_{i,k+1}^{(s)} \mid \theta\right).
\end{align}
For $S=1$, this reduces to the standard SGLD--Gibbs gradient
$G_k(\theta)$ defined in \cref{eq:sgld-gibbs-gradient}.
\cref{thm:joint_scaling_limit} generalizes to this case, with the only difference being 
that the diffusion matrix $A$ is replaced by
\begin{align}
A_S
&=
\frac{c_h}{c_\beta} \Gamma \mathbf{1}\{ \mathfrak{h}+\mathfrak{b}\leq \mathfrak{t} \}\\
&\phantom{=~} +
\frac{c_h^2}{4c_b}
\Gamma \widetilde I_\star^{(S)} \Gamma^\top
\mathbf{1}\{ \mathfrak{t} \leq \mathfrak{b}+\mathfrak{h} \}, 
\end{align}
where $I_\star^{(S)} \defas I_\star
+
\frac{1}{S}
M_\star$. 
Hence, we see that when $S > 1$, the ``Jensen gap'' $M_\star$ decreases to $M_\star / S$ and 
goes to zero as $S \to \infty$ (i.e., as the marginal likelihood estimate error goes to zero). 
However, smaller algorithm-induced  uncertainty comes at the cost of $S$ times as many gradient evaluations per iteration. 
Our result thus helps quantify the trade-off between computational cost
and accuracy of the uncertainty quantification.

\subsection{Parameter Tuning and uncertainty quantification}
To retain a nondegenerate and interpretable joint scaling limit, we
focus on the regime $\mathfrak h + \mathfrak b = 1$, in which the latent-variable
component admits a nontrivial macroscopic limit.
Our discussion covers general $S \ge 1$, with the special case $S=1$ recovering the standard SGLD--Gibbs update with a single conditional draw.
To keep the Gaussian noise active in the limit, we also set $\mathfrak t = 1$. 
Thus, we parameterize the tuning constants following the SGLD scaling-limit literature by setting
\[
\beta^{(n)} = \frac{n}{w_2},
\quad\text{and}\quad
h^{(n)} = \frac{4 w_1\, b^{(n)}}{n},
\]
where $w_1,w_2>0$ are constants. 
With these choices, the stationary covariance of the limiting OU process depends on
the choice of preconditioner $\Gamma$ and on $S$.
This leads to two natural tuning strategies.

\paragraph{Bernstein--von Mises covariance via $\Gamma = \bigl(\widetilde I_\star^{(S)}\bigr)^{-1}$.}
If we choose the preconditioner
$\Gamma = \bigl(\widetilde I_\star^{(S)}\bigr)^{-1}$,
then, by taking the weights to satisfy $w_1 + w_2 = 1$, the stationary covariance
is $J_\star^{-1}$,
recovering Bayes-type uncertainty quantification. 
Note that taking $w_1 = 0$ results in using SGD--Gibbs rather than SGLD--Gibbs. 

\paragraph{Bagged posterior and sandwich-type covariances via $\Gamma = J_\star^{-1}$.}
Alternatively, if we choose $\Gamma = J_\star^{-1}$,
the stationary covariance takes the bagged posterior form
\begin{align}
\lefteqn{w_1\, J_\star^{-1}\, \widetilde I_\star^{(S)}\, J_\star^{-1}
+
w_2\, J_\star^{-1}} \\
&\quad= 
w_1 J_\star^{-1} I_\star J_\star^{-1} + w_2 J_\star^{-1} + \dfrac{w_1}{S}\, J_\star^{-1} M_\star J_\star^{-1}.
\end{align}
The third term isolates the algorithm-induced contribution and decays as $S$ increases.
A sandwich-type covariance can thus be recovered by taking $w_1 = 1$ and $w_2 = 0$. 
The resulting tuning recommendations are summarized in \cref{tab:gibbs_sgld_tuning_recommendations}.

\begin{figure*}[!t]
  \centering
  \begin{minipage}{0.46\textwidth}
    \centering
    \includegraphics[width=0.9\linewidth]{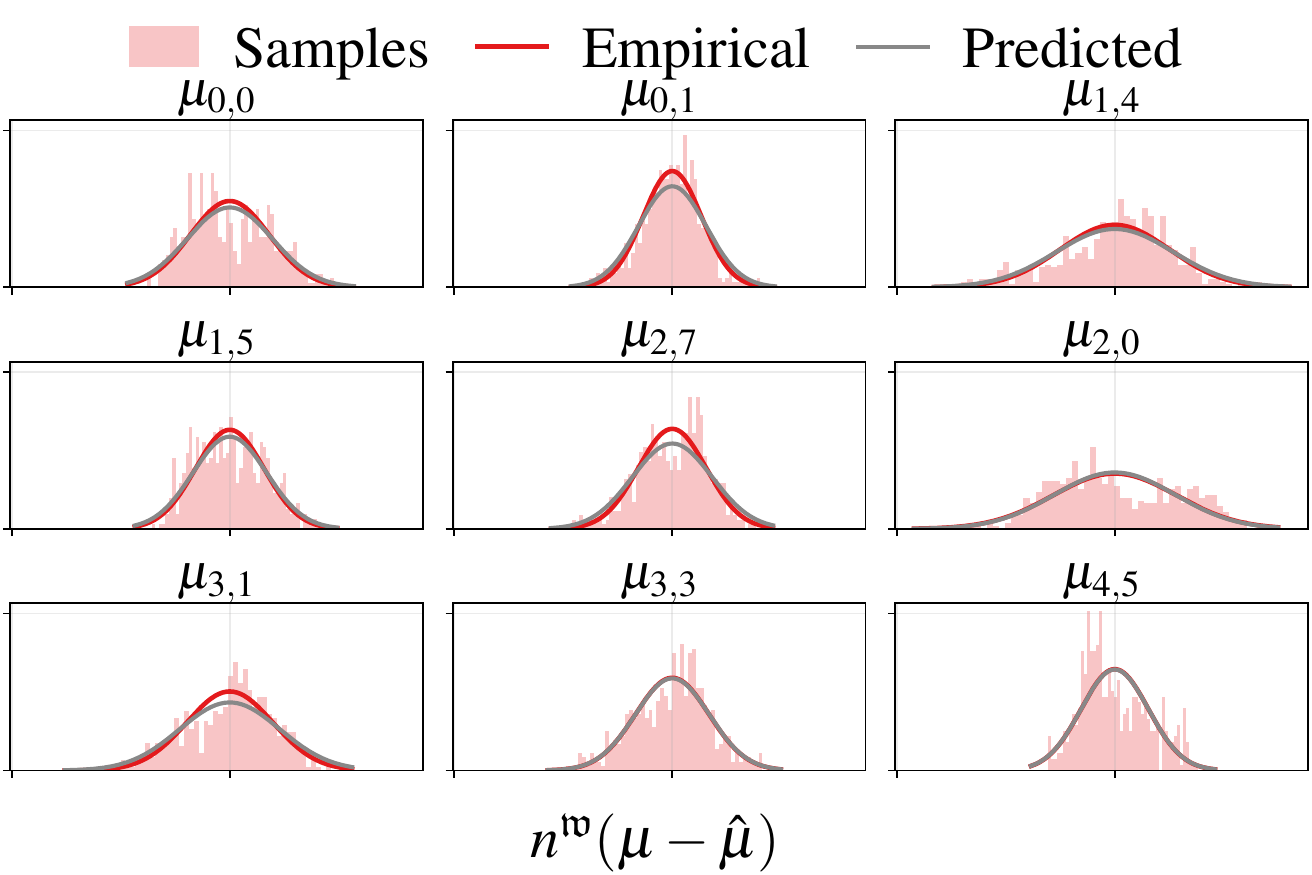}
    \caption*{
    \textbf{a.}
    Empirical stationary density of the rescaled global parameter $\vartheta^{(n)}$
    under SGLD--Gibbs, compared with the Ornstein--Uhlenbeck stationary distribution
    predicted by the scaling limit.
    }
  \end{minipage}
  \hfill
  \begin{minipage}{0.46\textwidth}
    \centering
    \includegraphics[width=0.9\linewidth]{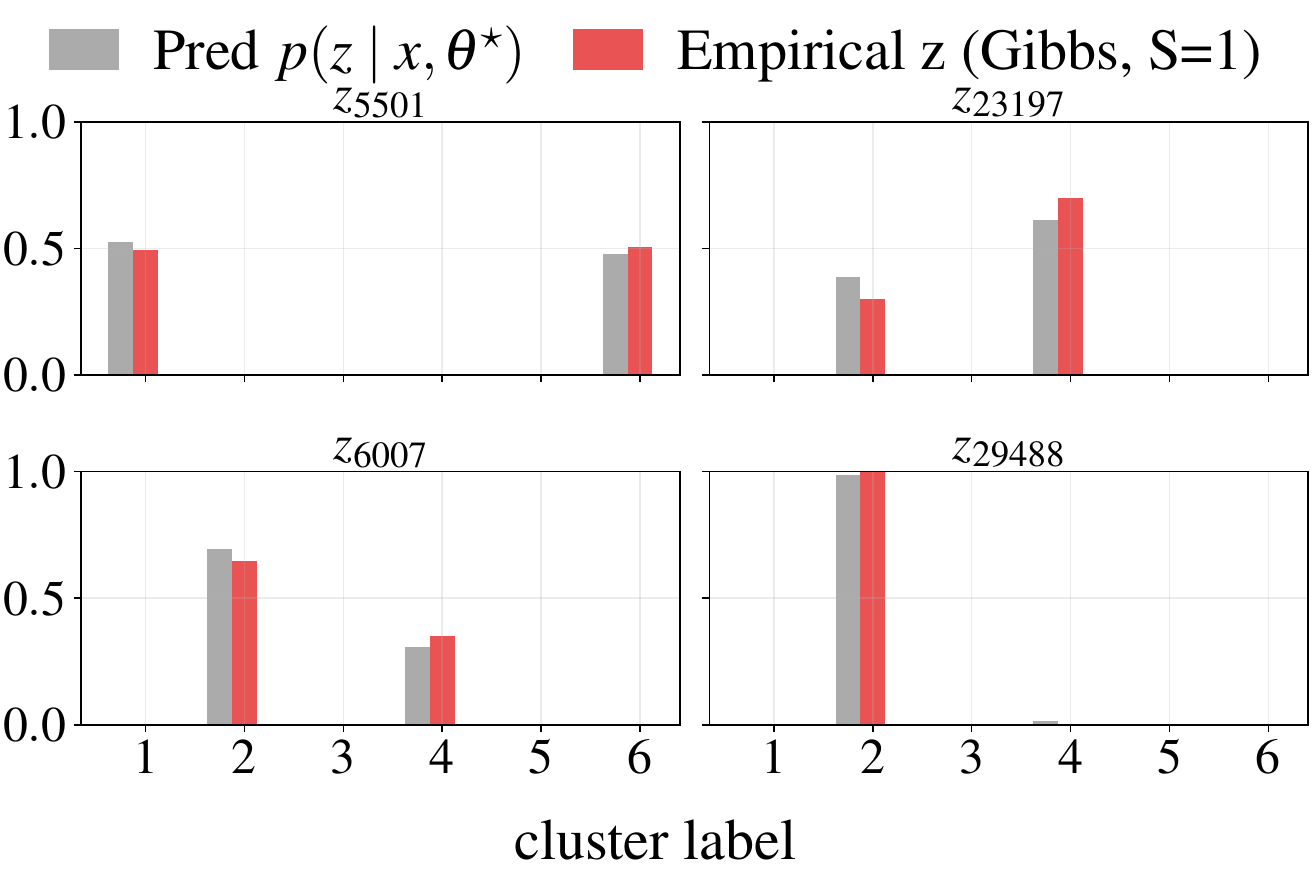}
    \caption*{
    \textbf{b.}
    Empirical stationary behavior of representative latent assignments,
    consistent with the jump-type limiting dynamics.
    }
  \end{minipage}

  \caption{
  Validation of the joint scaling limit on a synthetic Gaussian mixture model.
  }
  \label{fig:gmm-theory}
\end{figure*}

\subsection{Proof sketch of \cref{thm:joint_scaling_limit}}

Our proof follows the spirit of \citet{negrea2022statistical}. 
We establish weak convergence
of the processes in the Skorokhod topology in probability by first proving almost sure
convergence along subsequences. This is achieved by showing that the difference between
the approximate generator and the limiting generator, evaluated on smooth test functions
with compact support, vanishes uniformly. 
We divide the proof into two parts. 

In Part~1, we consider arguments that are sufficiently far from the support of the test
function.
The main idea is to control the probability that the global-parameter process
jumps back into the support of the test function.
A key difference with \citet{negrea2022statistical} is that we must impose assumptions on the
joint likelihood uniformly on latent variable values to ensure that, even in the presence of additional uncertainty induced by the
latent-variable updates, the probability that the global parameter jumps back into the
support can still be controlled at a comparable scale.

In Part~2, we analyze arguments that lie in or are close to the support of the
test function.
We perform a Taylor expansion of the joint approximate generator. 
The drift term converges in the same way as in
\citet{negrea2022statistical}. 
For the gradient component of the diffusion term, a similar
approach applies, but the resulting limit now explicitly incorporates additional variability arising
from sampling the latent variables. 
We then have to control a new term that bridges the infinitesimal operator
associated with Gibbs updates and the generator of a Poisson jump process. The most
technically challenging new aspect is to show that all cross terms involving both the
global parameters and the latent variables vanish in the limit. This vanishing further
implies that, in the asymptotic regime, the contribution of any single observation becomes
asymptotically independent of the global-parameter dynamics, which in turn allows the
joint limiting distribution to factorize into independent components.

\section{Experiments} \label{sec:experiments}

Our experiments are designed to answer three questions. 
First, we verify the predictions of our scaling-limit theory: under the hyperparameter tuning guided by \cref{tab:gibbs_sgld_tuning_recommendations}, samples produced by SGLD--Gibbs should match the stationary distributions predicted by the limiting processes. 
On synthetic models with known ground-truth parameters, we directly compare empirical posteriors of both global parameters and latent variables with their theoretical stationary distributions. 
Second, we evaluate the quality of uncertainty quantification provided by SGLD--Gibbs and compare it with stochastic variational inference (SVI), which we assess using posterior variances and rank-uniformity calibration diagnostics. 
Third, on real datasets where direct verification against ground-truth parameters is not possible, we evaluate downstream performance using clustering accuracy (ARI and AMI) for mixture models and held-out perplexity for topic models. Complete experimental details are provided in Appendix~\ref{app:exp}.

\subsection{Synthetic Gaussian Mixture Model}

\begin{figure*}[!t]
  \centering
  \begin{minipage}{0.46\textwidth}
    \centering
    \includegraphics[width=0.9\linewidth]{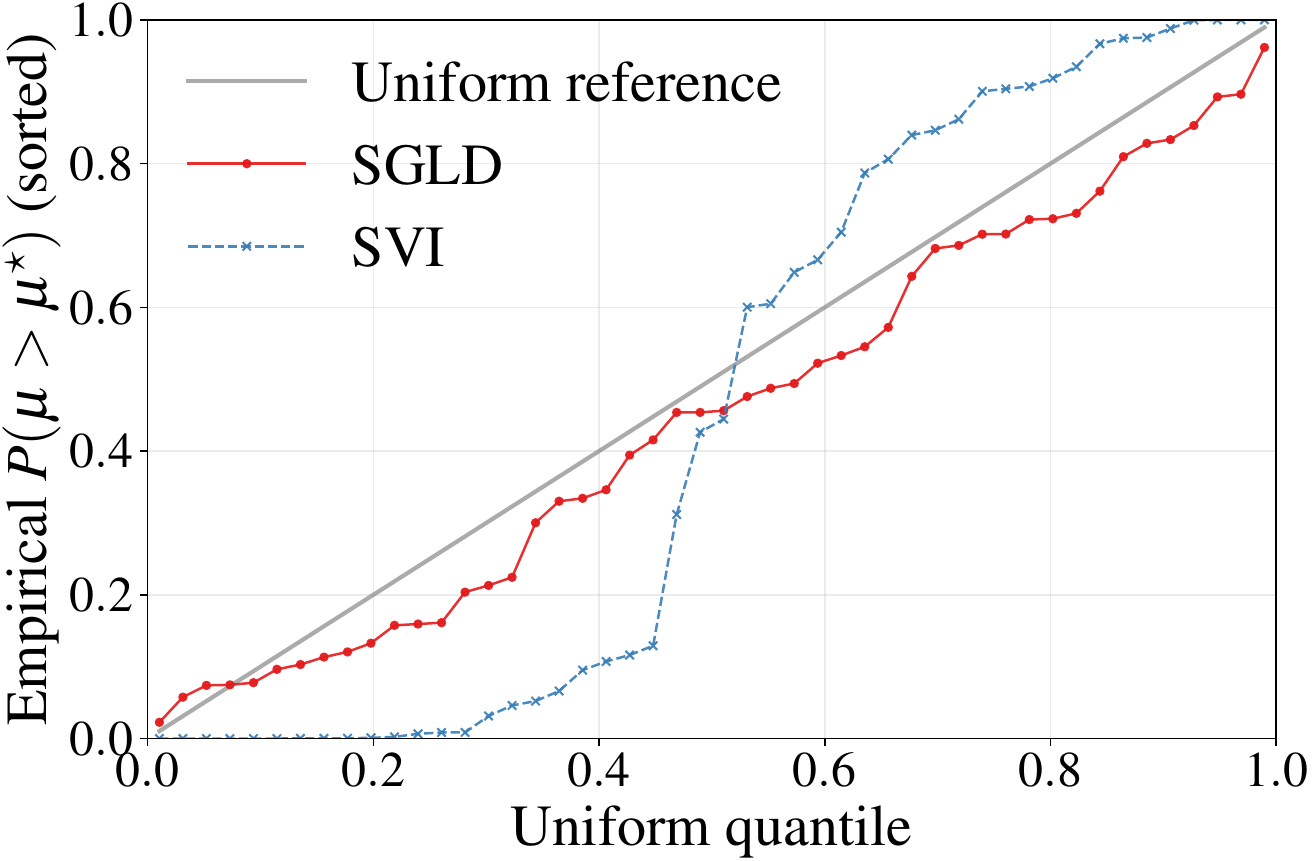}
    \caption*{
    \textbf{a.}
    Rank-uniformity diagnostic for global parameters.
    SGLD--Gibbs yields empirical ranks closer to the uniform reference line than
    stochastic variational inference (SVI), indicating better-calibrated uncertainty.
    }
  \end{minipage}
  \hfill
  \begin{minipage}{0.46\textwidth}
    \centering
    \includegraphics[width=0.9\linewidth]{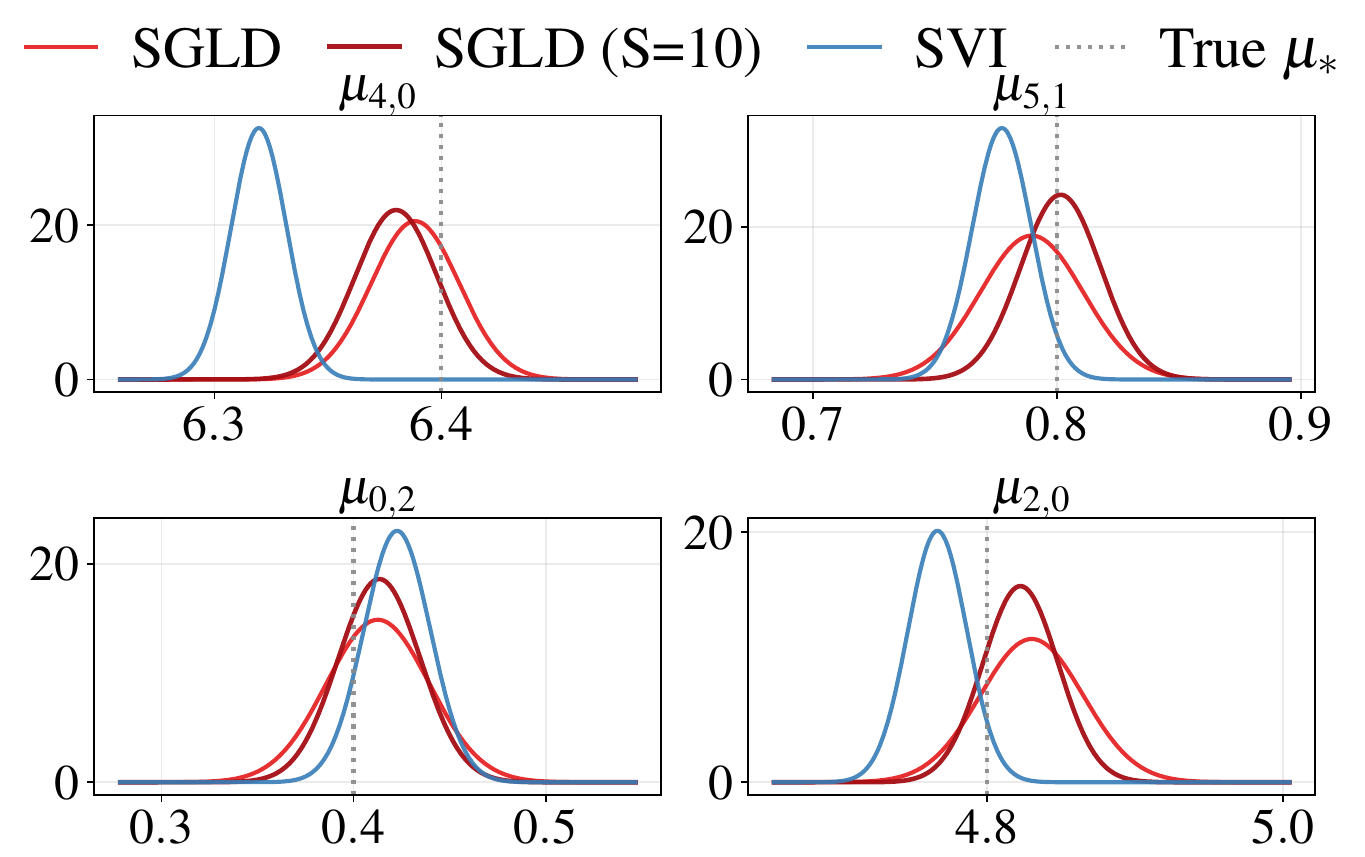}
    \caption*{
    \textbf{b.}
    Representative posterior marginals for selected parameters under SGLD--Gibbs and SVI,
    together with the true parameter values.
    SGLD--Gibbs concentrates more accurately around the ground truth. Increasing $S$ reduces uncertainty significantly.
    }
  \end{minipage}

  \caption{
  Synthetic Gaussian GMM: uncertainty quantification and posterior accuracy
  }
  \label{fig:gmm-uncertainty}
\end{figure*}

We generate synthetic data with sample size $n = 30{,}000$ with observations $x_i \in \mathbb{R}^8$.
The data are drawn from a finite Gaussian mixture with $6$ clusters.
We run SGLD--Gibbs using a minibatch size $b=50$ and consider Gibbs updates 
with $S \in \{1, 10\}$ samples. 
For SGLD, we follow the sandwich tuning and choose the step size and inverse temperature as $h = 2b/n$ and $\beta = 2n$.
We use a preconditioner for the global parameters constructed from an estimate of $J_\star^{-1}$, as described in \cref{tab:gibbs_sgld_tuning_recommendations}.
As a baseline, we apply SVI with a standard mean-field variational family.
We use a diagonal-covariance GMM with Normal--Gamma variational factors for the component means and precisions.

\Cref{fig:gmm-theory} demonstrates that our scaling limit theory accurately 
predicts the distributions of the global and latent variables. 
As shown in \cref{fig:gmm-uncertainty}(a),
SGLD--Gibbs also produces substantially better calibrated uncertainty than SVI, with empirical ranks lying close to the uniform reference.
Finally, \cref{fig:gmm-uncertainty}(b) demonstrates that SGLD--Gibbs also provides more accurate parameter estimates.
Moreover, as predicted by our theory, increasing the number of Gibbs draws to $S=10$ leads to a clear reduction in posterior uncertainty, resulting in visibly tighter marginal distributions compared to the $S=1$ case. 

\subsection{Synthetic Latent Dirichlet Allocation}
\begin{figure*}[!t]
  \centering
  \begin{minipage}{0.46\textwidth}
    \centering
    \includegraphics[width=0.9\linewidth]{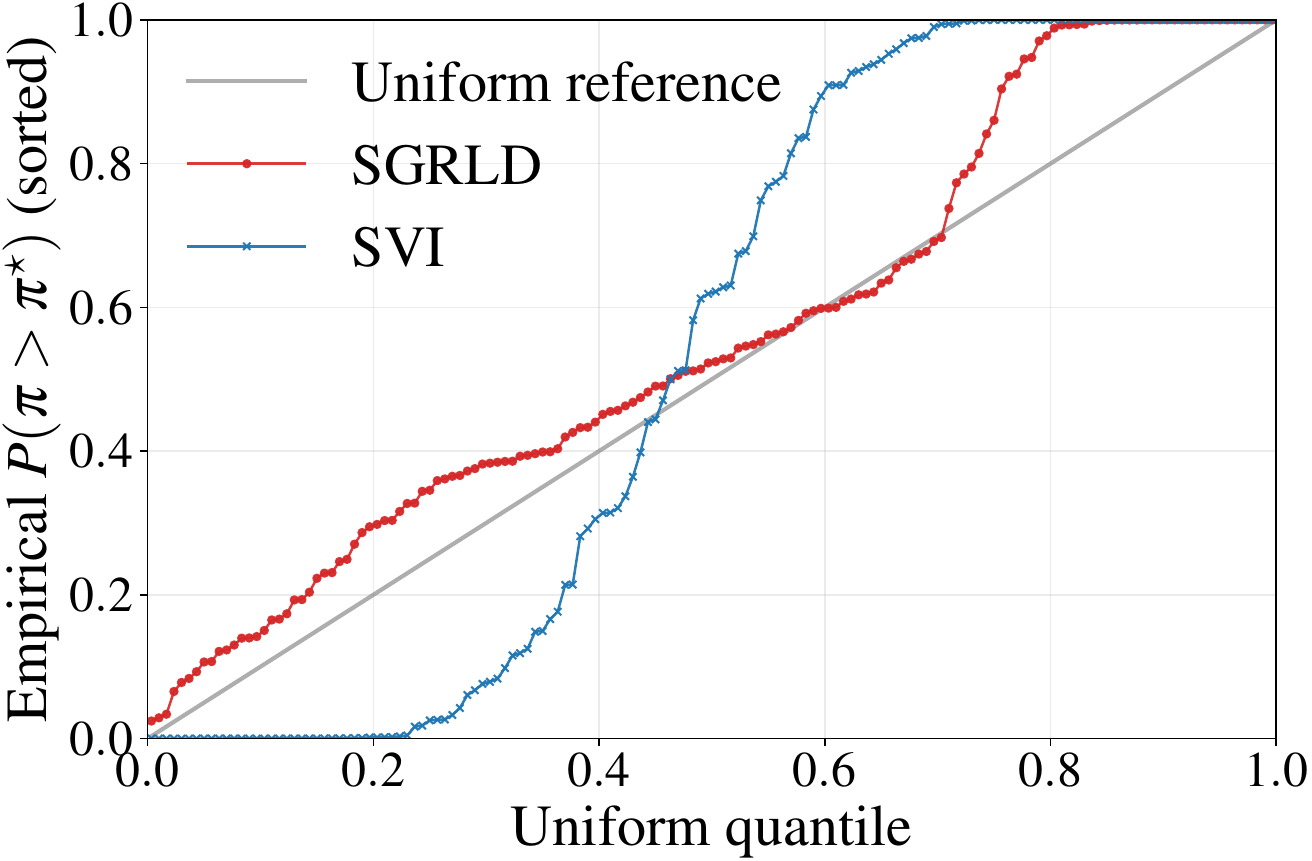}
    \caption*{
    \textbf{a.}
    Rank-uniformity diagnostic over topic-word probabilities.
    SGRLD--Gibbs yields empirical ranks closer to the uniform reference line,
    indicating better-calibrated uncertainty.
    }
  \end{minipage}
  \hfill
  \begin{minipage}{0.46\textwidth}
    \centering
    \includegraphics[width=0.9\linewidth]{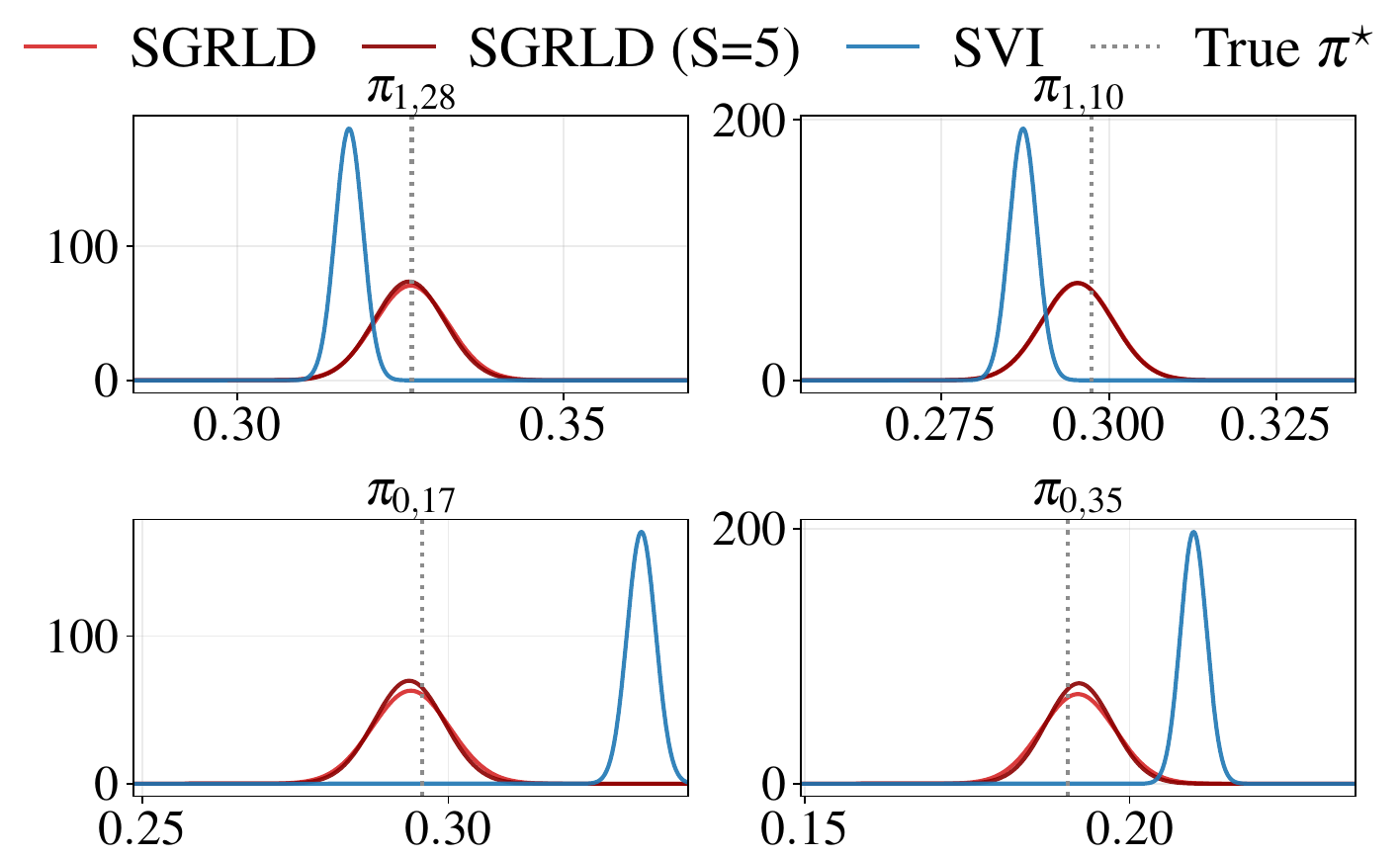}
    \caption*{
    \textbf{b.}
    Posterior marginals for selected topic-word probabilities under SGRLD--Gibbs and SVI,
    with true parameter values indicated.
    SGRLD--Gibbs concentrates more accurately around the ground truth with higher uncertainty. Increasing $S$ slightly reduces uncertainty.
    }
  \end{minipage}

  \caption{
  Synthetic LDA: uncertainty calibration and posterior accuracy.
  }
  \label{fig:lda-uncertainty}
\end{figure*}

We generate a synthetic topic model with vocabulary size $V=50$, number of topics $K=3$, and corpus size $d=10000$ documents.
Each document is generated from a Latent Dirichlet Allocation (LDA) model with fixed topic proportions and topic-word distributions.
We run SGLD--Gibbs using a minibatch size $b=100$ and consider Gibbs updates 
with $S \in \{1, 5\}$ samples. 
For SGLD we use the bagged posterior tuning
with $w_1 = w_2 = 1$ (so $h = 4b/n$ and $\beta = n$)
and for the preconditioner we follow \cite{LDA}.
This preconditioner is motivated by the natural geometry of the simplex, taking a diagonal form proportional to $\mathrm{diag}(\theta)$.
The resulting algorithm is known as \emph{stochastic gradient Riemannian Langevin dynamics} (SGRLD). 
As a baseline, we apply stochastic variational inference (SVI) to a semi-collapsed LDA model.
The document-level topic proportions are integrated out analytically, and we use a mean-field variational family in which the global topic-word distributions are modeled by Dirichlet factors and the latent topic assignments by categorical factors.

\cref{fig:lda-uncertainty}(a) 
shows that SGRLD--Gibbs yields ranks substantially closer to uniform than SVI, indicating superior calibration.
\cref{fig:lda-uncertainty}(b) 
shows that the SGRLD--Gibbs posterior concentrates more tightly around the ground-truth values and better matches the target uncertainty, whereas SVI exhibits noticeable miscalibration and bias in several coordinates. 
Using $S=5$ only slightly reduces uncertainty compared to $S=1$ (far less than in GMM), likely due to the small $S$ and the strong autocorrelation of Gibbs updates in LDA.

\subsection{Real-data evaluation: Flow Cytometry (GMM) and 20 Newsgroups (LDA)}
\label{sec:real_data}

\begin{table}[t]
  \caption{
  Performance comparison on real datasets.
  For Flow Cytometry (GMM), clustering quality is evaluated using
  adjusted Rand index (ARI) and adjusted mutual information (AMI).
  For 20 Newsgroups (LDA), predictive performance is measured by
  held-out perplexity (lower is better).
  }
  \label{tab:real_data_metrics}
  \begin{center}
    \begin{scriptsize}
      \begin{sc}
        \begin{tabular}{lccc}
          \toprule
          Dataset & Method & Metric & Result \\
          \midrule
          \multirow{2}{*}{Flow Cytometry}
            & SVI
            & ARI / AMI
            & $0.47 \; / \; 0.68$ \\
          & SGLD--Gibbs
            & ARI / AMI
            & $\mathbf{0.80} \; / \; \mathbf{0.82}$ \\
          \midrule
          \multirow{2}{*}{20 Newsgroups}
            & SVI
            & Perplexity
            & $2402$ \\
          & SGRLD--Gibbs
            & Perplexity
            & $\mathbf{2052}$ \\
          \bottomrule
        \end{tabular}
      \end{sc}
    \end{scriptsize}
  \end{center}
  \vskip -0.12in
  \vspace{-.5em}
\end{table}

We further evaluate SGLD--Gibbs on two real-world latent variable models:
a diagonal-covariance Gaussian mixture model (GMM) on a flow cytometry dataset
and latent Dirichlet allocation (LDA) on the 20 Newsgroups corpus.
\paragraph{Flow Cytometry (GMM).}
We measure inferential quality in terms of 
the adjusted Rand index (ARI) and adjusted mutual information (AMI), which compares the
inferred clustering to the provided expert clustering. 
\cref{tab:real_data_metrics} shows that SGLD--Gibbs attains substantially
higher ARI and AMI than SVI.
\paragraph{20 Newsgroups (LDA).}
For LDA, we compare methods based on predictive performance using held-out perplexity.
As shown in \cref{tab:real_data_metrics}, SGLD--Gibbs achieves much better perplexity ($\sim 350$ nats) than SVI.

Overall, these real-data results complement our synthetic experiments, demonstrating 
that in practice SGLD--Gibbs retains the scalability of SVI but 
provides superior task-specific performance.

\section{Discussion and Future Work}
This work provides a scaling-limit perspective on SGLD--Gibbs in latent variable
models, clarifying how uncertainty quantification and algorithmic tuning are shaped
by the interaction between stochastic-gradient dynamics and latent-variable
updates.
The resulting joint jump--diffusion limit explains how additional algorithm-induced uncertainty due to estimating the marginal likelihood with Gibbs samples 
contributes to the effective noise of the global parameters and yields
principled guidance for hyperparameter scaling.
Empirically, our results suggest that SGLD--Gibbs achieves better-calibrated
posterior uncertainty and better predictive performance than variational methods.

There are a number of limitations of our work that motivate directions for future research. 
Our results are for latent variable models in which each data point is
associated with a single local latent variable refreshed through Gibbs updates.
Many important models exhibit more complex dependency structures, such as Bayesian
matrix factorization, mixed-effects models, or hierarchical topic models, where
latent variables are shared across observations or interact globally.
Extending the joint scaling-limit framework to such settings would require
capturing richer coupling between latent variables and global parameters, and may
lead to qualitatively different limiting dynamics and hence distinct tuning guidance.
Moreover, when the latent-variable updates do not mix rapidly, for instance under
SGLD or Gibbs samplers with more complicated dependency structures, the resulting
latent variables can exhibit temporal dependence and nontrivial interactions
with the global-parameter iterates. Analyzing such regimes is an interesting
direction for future work.
\new{
Beyond these model-specific extensions, another important direction is to compare 
scaling-limit-based tuning with more algorithmic tuning criteria, such as the 
KSD-based strategy of \citet{10.1007/s11222-023-10233-3}.}

\section*{Impact Statement}
This paper presents work whose goal is to advance the field
of probabilistic machine learning. There are many potential
societal consequences of our work, none of which we feel must
be specifically highlighted here.
\bibliography{biblio}

\begin{thebibliography}{49}
\providecommand{\natexlab}[1]{#1}
\providecommand{\url}[1]{\texttt{#1}}
\expandafter\ifx\csname urlstyle\endcsname\relax
  \providecommand{\doi}[1]{doi: #1}\else
  \providecommand{\doi}{doi: \begingroup \urlstyle{rm}\Url}\fi

\bibitem[Ahn et~al.(2015)Ahn, Korattikara, Liu, Rajan, and Welling]{BayMatFac}
Ahn, S., Korattikara, A., Liu, N., Rajan, S., and Welling, M.
\newblock Large-scale distributed {B}ayesian matrix factorization using
  stochastic gradient {{MCMC}}, 2015.
\newblock URL \url{https://arxiv.org/abs/1503.01596}.

\bibitem[Anastasiou et~al.(2019)Anastasiou, Balasubramanian, and
  Erdogdu]{pmlr-v99-anastasiou19a}
Anastasiou, A., Balasubramanian, K., and Erdogdu, M.~A.
\newblock Normal approximation for stochastic gradient descent via
  non-asymptotic rates of martingale {CLT}.
\newblock In Beygelzimer, A. and Hsu, D. (eds.), \emph{Proceedings of the
  Thirty-Second Conference on Learning Theory}, volume~99 of \emph{Proceedings
  of Machine Learning Research}, pp.\  115--137. PMLR, 25--28 Jun 2019.
\newblock URL \url{https://proceedings.mlr.press/v99/anastasiou19a.html}.

\bibitem[Arous et~al.(2022)Arous, Gheissari, and Jagannath]{nips2022arous}
Arous, G.~B., Gheissari, R., and Jagannath, A.
\newblock High-dimensional limit theorems for {SGD}: effective dynamics and
  critical scaling.
\newblock In \emph{Proceedings of the 36th International Conference on Neural
  Information Processing Systems}, NIPS '22, Red Hook, NY, USA, 2022. Curran
  Associates Inc.
\newblock ISBN 9781713871088.

\bibitem[Bottou et~al.(2018)Bottou, Curtis, and Nocedal]{bottou2016}
Bottou, L., Curtis, F.~E., and Nocedal, J.
\newblock Optimization methods for large-scale machine learning.
\newblock \emph{SIAM Review}, 60\penalty0 (2):\penalty0 223--311, 2018.
\newblock \doi{10.1137/16M1080173}.
\newblock URL \url{https://doi.org/10.1137/16M1080173}.

\bibitem[Cheng et~al.(2020)Cheng, Yin, Bartlett, and
  Jordan]{cheng2020stochastic}
Cheng, X., Yin, D., Bartlett, P., and Jordan, M.
\newblock Stochastic gradient and {L}angevin processes.
\newblock In \emph{International Conference on Machine Learning}, pp.\
  1810--1819. PMLR, 2020.

\bibitem[Collins-Woodfin et~al.(2023)Collins-Woodfin, Paquette, Paquette, and
  Seroussi]{collinswoodfin2023hittinghighdimensionalnotesode}
Collins-Woodfin, E., Paquette, C., Paquette, E., and Seroussi, I.
\newblock Hitting the high-dimensional notes: An ode for sgd learning dynamics
  on glms and multi-index models, 2023.
\newblock URL \url{https://arxiv.org/abs/2308.08977}.

\bibitem[Collins-Woodfin et~al.(2024)Collins-Woodfin, Seroussi,
  Malaxechebarr\'{\i}a, Mackenzie, Paquette, and Paquette]{collins2024nips}
Collins-Woodfin, E., Seroussi, I., Malaxechebarr\'{\i}a, B.~n.~G., Mackenzie,
  A.~W., Paquette, E., and Paquette, C.
\newblock The high line: exact risk and learning rate curves of stochastic
  adaptive learning rate algorithms.
\newblock In \emph{Proceedings of the 38th International Conference on Neural
  Information Processing Systems}, NIPS '24, Red Hook, NY, USA, 2024. Curran
  Associates Inc.
\newblock ISBN 9798331314385.

\bibitem[Coullon et~al.(2023)Coullon, South, and
  Nemeth]{10.1007/s11222-023-10233-3}
Coullon, J., South, L., and Nemeth, C.
\newblock Efficient and generalizable tuning strategies for stochastic gradient
  mcmc.
\newblock \emph{Statistics and Computing}, 33\penalty0 (3), April 2023.
\newblock ISSN 0960-3174.
\newblock \doi{10.1007/s11222-023-10233-3}.
\newblock URL \url{https://doi.org/10.1007/s11222-023-10233-3}.

\bibitem[Danaher(2023)]{Peter2023OMA}
Danaher, P.~J.
\newblock Optimal microtargeting of advertising.
\newblock \emph{Journal of Marketing Research}, 60\penalty0 (3):\penalty0
  564--584, 2023.
\newblock \doi{10.1177/00222437221116034}.
\newblock URL \url{https://doi.org/10.1177/00222437221116034}.

\bibitem[Dieuleveut et~al.(2020)Dieuleveut, Durmus, and
  Bach]{10.1214/19-AOS1850}
Dieuleveut, A., Durmus, A., and Bach, F.
\newblock {Bridging the gap between constant step size stochastic gradient
  descent and {M}arkov chains}.
\newblock \emph{The Annals of Statistics}, 48\penalty0 (3):\penalty0 1348 --
  1382, 2020.
\newblock URL \url{https://doi.org/10.1214/19-AOS1850}.

\bibitem[Ethier \& Kurtz(2009)Ethier and Kurtz]{ethier2009markov}
Ethier, S.~N. and Kurtz, T.~G.
\newblock \emph{{Markov} Processes: Characterization and Convergence}.
\newblock Wiley Series in Probability and Statistics. John Wiley \& Sons, 2009.
\newblock ISBN 9780470412035.

\bibitem[Ge et~al.(2015)Ge, Huang, Jin, and Yuan]{ge2015escaping}
Ge, R., Huang, F., Jin, C., and Yuan, Y.
\newblock Escaping from saddle points—online stochastic gradient for tensor
  decomposition.
\newblock In \emph{Conference on learning theory}, pp.\  797--842. PMLR, 2015.

\bibitem[Gelman et~al.(2013)Gelman, Carlin, Stern, Dunson, Vehtari, and
  Rubin]{gelman2013bayesian}
Gelman, A., Carlin, J., Stern, H., Dunson, D., Vehtari, A., and Rubin, D.
\newblock \emph{{B}ayesian Data Analysis}.
\newblock Chapman \& Hall/CRC Texts in Statistical Science Series. CRC, Boca
  Raton, Florida, third edition, 2013.
\newblock ISBN 9781439840955 1439840954.
\newblock URL \url{https://stat.columbia.edu/~gelman/book/}.

\bibitem[Giordano et~al.(2018)Giordano, Broderick, and
  Jordan]{JASA2018giordano}
Giordano, R., Broderick, T., and Jordan, M.~I.
\newblock Covariances, robustness and {V}ariational {B}ayes.
\newblock \emph{J. Mach. Learn. Res.}, 19\penalty0 (1):\penalty0 1981–2029,
  January 2018.
\newblock ISSN 1532-4435.

\bibitem[Hoffman et~al.(2010)Hoffman, Bach, and Blei]{NIPS2010_OVB}
Hoffman, M., Bach, F., and Blei, D.
\newblock Online learning for {L}atent {D}irichlet {A}llocation.
\newblock In Lafferty, J., Williams, C., Shawe-Taylor, J., Zemel, R., and
  Culotta, A. (eds.), \emph{Advances in Neural Information Processing Systems},
  volume~23. Curran Associates, Inc., 2010.
\newblock URL
  \url{https://proceedings.neurips.cc/paper_files/paper/2010/file/71f6278d140af599e06ad9bf1ba03cb0-Paper.pdf}.

\bibitem[Hoffman et~al.(2013)Hoffman, Blei, Wang, and
  Paisley]{JMLR:v14:hoffman13a}
Hoffman, M.~D., Blei, D.~M., Wang, C., and Paisley, J.
\newblock Stochastic variational inference.
\newblock \emph{Journal of Machine Learning Research}, 14\penalty0
  (40):\penalty0 1303--1347, 2013.
\newblock URL \url{http://jmlr.org/papers/v14/hoffman13a.html}.

\bibitem[Huggins \& Miller(2024)Huggins and Miller]{huggins2024reproducible}
Huggins, J.~H. and Miller, J.~W.
\newblock {Reproducible parameter inference using bagged posteriors}.
\newblock \emph{Electronic Journal of Statistics}, 18\penalty0 (1), 2024.
\newblock ISSN 1935-7524.
\newblock \doi{10.1214/24-ejs2237}.

\bibitem[Jin et~al.(2017)Jin, Ge, Netrapalli, Kakade, and
  Jordan]{jin2017escape}
Jin, C., Ge, R., Netrapalli, P., Kakade, S.~M., and Jordan, M.~I.
\newblock How to escape saddle points efficiently.
\newblock In \emph{International conference on machine learning}, pp.\
  1724--1732. PMLR, 2017.

\bibitem[Kleijn \& van~der Vaart(2012)Kleijn and van~der
  Vaart]{kleijn2012bernstein}
Kleijn, B. and van~der Vaart, A.
\newblock The {Bernstein-Von-Mises} theorem under misspecification.
\newblock \emph{Electronic Journal of Statistics}, \penalty0 (6):\penalty0
  354--381, 2012.

\bibitem[Kucukelbir et~al.(2017)Kucukelbir, Tran, Ranganath, Gelman, and
  Blei]{ADVI2017}
Kucukelbir, A., Tran, D., Ranganath, R., Gelman, A., and Blei, D.~M.
\newblock Automatic differentiation variational inference.
\newblock \emph{J. Mach. Learn. Res.}, 18\penalty0 (1):\penalty0 430–474,
  January 2017.
\newblock ISSN 1532-4435.

\bibitem[Kushner \& Yin(2003)Kushner and Yin]{kushner2003stochastic}
Kushner, H. and Yin, G.
\newblock \emph{Stochastic Approximation and Recursive Algorithms and
  Applications}.
\newblock Stochastic Modelling and Applied Probability. Springer New York,
  2003.
\newblock ISBN 9780387008943.
\newblock URL \url{https://books.google.com/books?id=_0bIieuUJGkC}.

\bibitem[Kushner \& Huang(1981)Kushner and Huang]{kushner1981}
Kushner, H.~J. and Huang, H.
\newblock Asymptotic properties of stochastic approximations with constant
  coefficients.
\newblock \emph{SIAM Journal on Control and Optimization}, 19\penalty0
  (1):\penalty0 87--105, 1981.
\newblock \doi{10.1137/0319007}.
\newblock URL \url{https://doi.org/10.1137/0319007}.

\bibitem[Kushner \& Yang(1993)Kushner and Yang]{kushner1993}
Kushner, H.~J. and Yang, J.
\newblock Stochastic approximation with averaging of the iterates: Optimal
  asymptotic rate of convergence for general processes.
\newblock \emph{SIAM Journal on Control and Optimization}, 31\penalty0
  (4):\penalty0 1045--1062, 1993.
\newblock \doi{10.1137/0331047}.
\newblock URL \url{https://doi.org/10.1137/0331047}.

\bibitem[Li et~al.(2016)Li, Ahn, and Welling]{MMSB}
Li, W., Ahn, S., and Welling, M.
\newblock Scalable {MCMC} for mixed membership stochastic blockmodels.
\newblock In Gretton, A. and Robert, C.~C. (eds.), \emph{Proceedings of the
  19th International Conference on Artificial Intelligence and Statistics},
  volume~51 of \emph{Proceedings of Machine Learning Research}, pp.\  723--731,
  Cadiz, Spain, 09--11 May 2016. PMLR.
\newblock URL \url{https://proceedings.mlr.press/v51/li16d.html}.

\bibitem[Loaiza-Maya \& Nibbering(2023)Loaiza-Maya and
  Nibbering]{Loaiza-Maya02102023}
Loaiza-Maya, R. and Nibbering, D.
\newblock Fast {V}ariational {B}ayes methods for multinomial probit models.
\newblock \emph{Journal of Business \& Economic Statistics}, 41\penalty0
  (4):\penalty0 1352--1363, 2023.
\newblock \doi{10.1080/07350015.2022.2139267}.
\newblock URL \url{https://doi.org/10.1080/07350015.2022.2139267}.

\bibitem[Loaiza-Maya et~al.(2024)Loaiza-Maya, Nibbering, and
  Zhu]{LOAIZAMAYA2024105741}
Loaiza-Maya, R., Nibbering, D., and Zhu, D.
\newblock Hybrid unadjusted {L}angevin methods for high-dimensional latent
  variable models.
\newblock \emph{Journal of Econometrics}, 241\penalty0 (2):\penalty0 105741,
  2024.
\newblock ISSN 0304-4076.
\newblock \doi{https://doi.org/10.1016/j.jeconom.2024.105741}.
\newblock URL
  \url{https://www.sciencedirect.com/science/article/pii/S0304407624000873}.

\bibitem[Mandt et~al.(2017)Mandt, Hoffman, and Blei]{mandt2017stochastic}
Mandt, S., Hoffman, M.~D., and Blei, D.~M.
\newblock Stochastic gradient descent as approximate {B}ayesian inference.
\newblock \emph{Journal of Machine Learning Research}, 18\penalty0
  (134):\penalty0 1--35, 2017.
\newblock URL \url{http://jmlr.org/papers/v18/17-214.html}.

\bibitem[Margossian et~al.(2025)Margossian, Pillaud-Vivien, and
  Saul]{margossian2025variational}
Margossian, C.~C., Pillaud-Vivien, L., and Saul, L.~K.
\newblock Variational inference for uncertainty quantification: An analysis of
  trade-offs.
\newblock \emph{Journal of Machine Learning Research}, 26\penalty0
  (202):\penalty0 1--41, 2025.

\bibitem[Mcleish(1976)]{Mcleish1976}
Mcleish, D.~L.
\newblock Functional and random central limit theorems for the
  {R}obbins-{M}unro process, 1976.
\newblock URL \url{https://www.jstor.org/stable/3212676}.

\bibitem[Mignacco et~al.(2021)Mignacco, Krzakala, Urbani, and
  Zdeborová]{Mignacco_2021}
Mignacco, F., Krzakala, F., Urbani, P., and Zdeborová, L.
\newblock Dynamical mean-field theory for stochastic gradient descent in
  {G}aussian mixture classification*.
\newblock \emph{Journal of Statistical Mechanics: Theory and Experiment},
  2021\penalty0 (12):\penalty0 124008, December 2021.
\newblock ISSN 1742-5468.
\newblock \doi{10.1088/1742-5468/ac3a80}.
\newblock URL \url{http://dx.doi.org/10.1088/1742-5468/ac3a80}.

\bibitem[Mou et~al.(2020)Mou, Li, Wainwright, Bartlett, and
  Jordan]{pmlr-v125-mou20a}
Mou, W., Li, C.~J., Wainwright, M.~J., Bartlett, P.~L., and Jordan, M.~I.
\newblock On linear stochastic approximation: Fine-grained {P}olyak-{R}uppert
  and non-asymptotic concentration.
\newblock In Abernethy, J. and Agarwal, S. (eds.), \emph{Proceedings of Thirty
  Third Conference on Learning Theory}, volume 125 of \emph{Proceedings of
  Machine Learning Research}, pp.\  2947--2997. PMLR, 09--12 Jul 2020.
\newblock URL \url{https://proceedings.mlr.press/v125/mou20a.html}.

\bibitem[Moulines \& Bach(2011)Moulines and Bach]{NIPS2011_40008b9a}
Moulines, E. and Bach, F.
\newblock Non-asymptotic analysis of stochastic approximation algorithms for
  machine learning.
\newblock In Shawe-Taylor, J., Zemel, R., Bartlett, P., Pereira, F., and
  Weinberger, K. (eds.), \emph{Advances in Neural Information Processing
  Systems}, volume~24. Curran Associates, Inc., 2011.
\newblock URL
  \url{https://proceedings.neurips.cc/paper_files/paper/2011/file/40008b9a5380fcacce3976bf7c08af5b-Paper.pdf}.

\bibitem[Murphy(2023)]{pml2Book}
Murphy, K.~P.
\newblock \emph{Probabilistic Machine Learning: Advanced Topics}.
\newblock MIT Press, 2023.
\newblock URL \url{http://probml.github.io/book2}.

\bibitem[Negrea et~al.(2023)Negrea, Yang, Feng, Roy, and
  Huggins]{negrea2022statistical}
Negrea, J., Yang, J., Feng, H., Roy, D.~M., and Huggins, J.~H.
\newblock Tuning stochastic gradient algorithms for statistical inference via
  large-sample asymptotics, 2023.
\newblock URL \url{https://arxiv.org/abs/2207.12395}.

\bibitem[Nemeth \& Fearnhead(2021)Nemeth and Fearnhead]{sgmcmc2021}
Nemeth, C. and Fearnhead, P.
\newblock Stochastic gradient {Markov} chain {M}onte {C}arlo.
\newblock \emph{Journal of the American Statistical Association}, 116\penalty0
  (533):\penalty0 433--450, 2021.
\newblock \doi{10.1080/01621459.2020.1847120}.
\newblock URL \url{https://doi.org/10.1080/01621459.2020.1847120}.

\bibitem[Nemirovski et~al.(2009)Nemirovski, Juditsky, Lan, and
  Shapiro]{Nemirovski2009}
Nemirovski, A., Juditsky, A., Lan, G., and Shapiro, A.
\newblock Robust stochastic approximation approach to stochastic programming.
\newblock \emph{SIAM Journal on Optimization}, 19\penalty0 (4):\penalty0
  1574--1609, 2009.
\newblock URL \url{https://doi.org/10.1137/070704277}.

\bibitem[Patterson \& Teh(2013)Patterson and Teh]{LDA}
Patterson, S. and Teh, Y.~W.
\newblock Stochastic gradient {R}iemannian {L}angevin dynamics on the
  probability simplex.
\newblock In Burges, C., Bottou, L., Welling, M., Ghahramani, Z., and
  Weinberger, K. (eds.), \emph{Advances in Neural Information Processing
  Systems}, volume~26. Curran Associates, Inc., 2013.
\newblock URL
  \url{https://proceedings.neurips.cc/paper/2013/file/309928d4b100a5d75adff48a9bfc1ddb-Paper.pdf}.

\bibitem[Pedregosa et~al.(2011)Pedregosa, Varoquaux, Gramfort, Michel, Thirion,
  Grisel, Blondel, Prettenhofer, Weiss, Dubourg, et~al.]{scikit-learn}
Pedregosa, F., Varoquaux, G., Gramfort, A., Michel, V., Thirion, B., Grisel,
  O., Blondel, M., Prettenhofer, P., Weiss, R., Dubourg, V., et~al.
\newblock Scikit-learn: Machine learning in python.
\newblock \emph{Journal of Machine Learning Research}, 12:\penalty0 2825--2830,
  2011.

\bibitem[Pflug(1986)]{pflug1986}
Pflug, G.~C.
\newblock Stochastic minimization with constant step-size: Asymptotic laws.
\newblock \emph{SIAM Journal on Control and Optimization}, 24\penalty0
  (4):\penalty0 655--666, 1986.
\newblock \doi{10.1137/0324039}.
\newblock URL \url{https://doi.org/10.1137/0324039}.

\bibitem[Polyak \& Juditsky(1992)Polyak and Juditsky]{doi:10.1137/0330046}
Polyak, B.~T. and Juditsky, A.~B.
\newblock Acceleration of stochastic approximation by averaging.
\newblock \emph{SIAM Journal on Control and Optimization}, 30\penalty0
  (4):\penalty0 838--855, 1992.
\newblock URL \url{https://doi.org/10.1137/0330046}.

\bibitem[Qian et~al.(2024)Qian, Xie, Liu, and
  Zhang]{qian2024sureconvergenceratesconcentration}
Qian, X., Xie, Z., Liu, X., and Zhang, S.
\newblock Almost sure convergence rates and concentration of stochastic
  approximation and reinforcement learning with {M}arkovian noise, 2024.
\newblock URL \url{https://arxiv.org/abs/2411.13711}.

\bibitem[Rakhlin et~al.(2011)Rakhlin, Shamir, and Sridharan]{rakhlin2011making}
Rakhlin, A., Shamir, O., and Sridharan, K.
\newblock Making gradient descent optimal for strongly convex stochastic
  optimization.
\newblock \emph{arXiv preprint arXiv:1109.5647}, 2011.

\bibitem[Ruppert(1988)]{ruppert1988}
Ruppert, D.
\newblock Efficient estimations from a slowly convergent {R}obbins-{M}onro
  process.
\newblock 02 1988.

\bibitem[Srikant(2024)]{srikant2024rates}
Srikant, R.
\newblock Rates of convergence in the central limit theorem for {M}arkov
  chains, with an application to td learning, 2024.

\bibitem[Walk(1977)]{Walk1977AnIP}
Walk, H.
\newblock An invariance principle for the {R}obbins-{M}onro process in a
  {H}ilbert space.
\newblock \emph{Zeitschrift f{\"u}r Wahrscheinlichkeitstheorie und Verwandte
  Gebiete}, 39:\penalty0 135--150, 1977.
\newblock URL \url{https://api.semanticscholar.org/CorpusID:119733417}.

\bibitem[Wang et~al.(2025)Wang, Kasprzak, Negrea, Bourguin, and
  Huggins]{wang2025quantitativeerrorboundsscaling}
Wang, X., Kasprzak, M.~J., Negrea, J., Bourguin, S., and Huggins, J.~H.
\newblock Quantitative error bounds for scaling limits of stochastic iterative
  algorithms, 2025.
\newblock URL \url{https://arxiv.org/abs/2501.12212}.

\bibitem[Wang et~al.(2026)Wang, Ding, and Huggins]{wang2026accurate}
Wang, Y., Ding, J., and Huggins, J.~H.
\newblock Accurate large-sample uncertainty quantification using stochastic
  gradient {M}arkov chain {M}onte {C}arlo.
\newblock In \emph{Proceedings of the 43rd International Conference on Machine
  Learning}, Proceedings of Machine Learning Research. PMLR, 2026.

\bibitem[Welling \& Teh(2011)Welling and Teh]{conf/icml/WellingT11}
Welling, M. and Teh, Y.~W.
\newblock {B}ayesian learning via stochastic gradient {L}angevin dynamics.
\newblock In Getoor, L. and Scheffer, T. (eds.), \emph{ICML}, pp.\  681--688.
  Omnipress, 2011.
\newblock URL
  \url{http://dblp.uni-trier.de/db/conf/icml/icml2011.html#WellingT11}.

\bibitem[White(1982)]{white1982maximum}
White, H.
\newblock Maximum likelihood estimation of misspecified models.
\newblock \emph{Econometrica: Journal of the econometric society}, pp.\  1--25,
  1982.

\end{thebibliography}
\bibliographystyle{icml2026}

\appendix
\onecolumn

\counterwithin{equation}{section}
\counterwithin{figure}{section}
\renewcommand{\thefigure}{\Alph{section}.\arabic{figure}}
\counterwithin{table}{section}
\renewcommand{\thetable}{\Alph{section}.\arabic{table}}
\counterwithin{theorem}{section}
\renewcommand{\thetheorem}{\Alph{section}.\arabic{theorem}}
\renewcommand{\thelemma}{\Alph{section}.\arabic{lemma}}
\renewcommand{\theassumption}{\Alph{section}.\arabic{assumption}}
\renewcommand{\thedefinition}{\Alph{section}.\arabic{definition}}
\renewcommand{\theproposition}{\Alph{section}.\arabic{proposition}}

\section{Preliminaries for Proof of Main Result}

\subsection{Assumptions}
\label{app:assumptions}
Recall that we assume throughout that $\obs{\obsidx} \sim P$ independently for all $\obsidx \in \mathbb{N}$. We denote
$\ell(\globpar;\obs,z) \defas \log p(\obs,z \mid \globpar)$.

\begin{assumption}
\label{Asp1}
$ \nabla \log \prior$ is $L_0$-Lipschitz, and $\log p(x,z\mid \cdot) \in C^2(\Theta)$ for each $x,z \in (\obsspace,\mathbb{R}^m)$
\end{assumption}

\begin{assumption}
The exponents satisfy $\powerstep-\powerspa-\powertime/3> 0$ and $\expect [\|\nabla \log p(\obs{1},\cdot\mid \globpar^*)\|_{\infty}^{p_{2}}] < \infty$ for some $p_{2} >\frac{1}{\powerstep-\powerspa-\powertime/3}$. 
\end{assumption}
\begin{assumption}
For some $q_3 \in [0,\powerspa]$ and $p_3:=\frac{1}{\powerstep+q_{3}-\powerspa-\powertime/3}$, the local critical points satisfy $\|\mleglobno -\globpar^*\| \in o_{p}(1/n^{q_3})$, and 
$\expect[\|\nabla^{\otimes 2} \log  (\obs{1},\cdot\mid \cdot)\|^{p_3}] < \infty$. 
\end{assumption}

Let 
\[
\ell(\theta; x_i) := \log p(x_i \mid \theta) = \log \int p(x_i, z_i \mid \theta) \, dz_i
\]
denote the log-likelihood function with the latent variable \( z_i \) marginalized out. For any \( r > 0 \), let 
\[
B^{(n)}(r) := B(\hat{\theta}_n, r / n^\powerspa)
\]
denote the ball centered at the MLE \( \hat{\theta}_n \) with radius \( r / n^\powerspa \), for some scaling exponent \( \alpha > 0 \).
\begin{assumption}
 There is a non-decreasing sequence $r_{J,n}\to \infty$ such that
\[\sup_{\globpar\in B(r_{J,n})}\|\frac{1}{n}\sum_{\obsidx=1}^{\subseq}\nabla_{\globpar}^{\otimes 2}\ell(\globpar;\obs{\obsidx})+ J_{*}\| \to 0\]
\end{assumption}

\begin{assumption}
There is a non-decreasing sequence $r_{I,n}\to \infty$ such that
\[\sup_{\globpar\in B(r_{I,n})}\|\frac{1}{n}\sum_{\obsidx=1}^{\subseq}\expect_{z|\globpar,\ob}[\nabla_{\globpar}\log p(\obs{\obsidx},z_{\obsidx}\mid \globpar)^{\otimes 2}]- \tilde I_\star\| \to 0\]
\end{assumption}
\begin{assumption}
\label{asp6}
If  $\globpar_n \to \globpar^*$ when $n\to \infty$, then
for almost every $X$ and $z$, 
\[p(z|X,\globpar_n) \to p(z|X,\globpar^*) \]
when $n \to \infty.$

\end{assumption}

\subsection{Technical Lemmas}

We will make use of the following two technical results in our proof. 

\begin{proposition}[Approximation of Markov chains \citep{ethier2009markov}]
\label{prop:markov_chain_approximation}
Let
\[
A : C_c^\infty(\mathbb{R}^d) \to C(\mathbb{R}^d)
\]
be a linear operator, and suppose that the closure of the graph of $A$
with respect to the graph norm
\[
\|f\|_A := \|f\|_\infty + \|Af\|_\infty,
\qquad f \in C_c^\infty(\mathbb{R}^d),
\]
generates a Feller semigroup $(T_t)_{t\ge 0}$ on $\mathbb{R}^d$.
Let $(\theta_t)_{t\ge 0}$ be a Markov process with forward operator semigroup
$(T_t)_{t\ge 0}$.
Let $\{(\theta^{(n)}_k)_{k\in\mathbb{N}\cup\{0\}}\}_{n\in\mathbb{N}}$
be a sequence of discrete-time Markov chains on $\mathbb{R}^d$
with respective transition kernels $\{U^{(n)}\}_{n\in\mathbb{N}}$.
Suppose that $0 < \alpha^{(n)} \to \infty$, and define
\[
A^{(n)} := \alpha^{(n)} \bigl(U^{(n)} - I\bigr),
\qquad
T_t^{(n)} := \bigl(U^{(n)}\bigr)^{\lfloor \alpha^{(n)} t \rfloor},
\qquad
\theta_t^{(n)} := \theta^{(n)}_{\lfloor \alpha^{(n)} t \rfloor}.
\]
If
\[
\|A^{(n)} f - A f\|_\infty \;\longrightarrow\; 0
\quad\text{for all } f \in C_c^\infty(\mathbb{R}^d),
\]
then
\begin{enumerate}
\item[(a)]
$T_t^{(n)} \to T_t$ for each $t>0$ and 
\item[(b)]
if $\theta^{(n)}(0) \Rightarrow \theta(0)$, then
\[
\theta^{(n)}(\cdot) \Rightarrow \theta(\cdot)
\quad\text{in the Skorokhod topology}.
\]
\end{enumerate}
\end{proposition}

\begin{lemma}[\citet{negrea2022statistical}]
\label{lem:subsequence_convergence}
Let $(\Omega,\mathcal{F},\mathbb{P})$ be a probability space, let $(\mathcal{X},\tau)$
be a topological space endowed with the $\sigma$-field
$\mathcal{F}_{\mathcal{X}} := \sigma(\tau)$,
let $(X_n)_{n\in\mathbb{N}}$ be a sequence of $\mathcal{X}$-valued random elements,
and let $x \in \mathcal{X}$.
If for every subsequence $(n_m)$ there exists a sub-subsequence $(n_{m_k})$
such that
\[
X_{n_{m_k}} \;\longrightarrow\; x
\quad\text{almost surely as } k \to \infty,
\]
then
\[
X_n \;\xrightarrow{\;\mathbb{P}\;}\; x.
\]
If $(\mathcal{X},\tau)$ is first countable, then the converse also holds:
if
\[
X_n \;\xrightarrow{\;\mathbb{P}\;}\; x,
\]
then for every subsequence $(n_m)$ there exists a sub-subsequence $(n_{m_k})$
such that
\[
X_{n_{m_k}} \;\longrightarrow\; x
\quad\text{almost surely as } k \to \infty.
\]
\end{lemma}

\subsection{Reduction to almost-sure convergence along subsequences}

Define the random quantities
\[
\begin{aligned}
\Phi^{(n)}
&:= \max\!\left\{\Phi^{(n)}_1,\Phi^{(n)}_2,\Phi^{(n)}_3\right\},\\
\Phi^{(n)}_1
&:= n^{q_3}\,\|\hat\theta^{(n)}-\theta^\star\|,\\
\Phi^{(n)}_2
&:= \sup_{\theta\in B(r_{J,n})}
\left\|
\frac{1}{n}\sum_{i=1}^n
\nabla_\theta^{\otimes 2}\ell(\theta;X_i)
+ J_\star
\right\|,\\
\Phi^{(n)}_3
&:= \sup_{\theta\in B(r_{I,n})}
\left\|
\frac{1}{n}\sum_{i=1}^n
\mathbb{E}_{Z_i\mid X_i,\theta}
\!\left[
\nabla_\theta \log p(X_i,Z_i\mid\theta)^{\otimes 2}
\right]
- \tilde I_\star
\right\|.
\end{aligned}
\]

By assumption, $\Phi^{(n)} \xrightarrow{\mathbb{P}} 0$.
By Lemma~1 of \citet{negrea2022statistical}, for every subsequence
$(n_m)_{m\in\mathbb{N}}$ there exists a further subsequence
$(n_{m_k})_{k\in\mathbb{N}}$ such that
\[
\Phi^{(n_{m_k})} \xrightarrow{\mathrm{a.s.}} 0 .
\]

It therefore suffices to establish weak convergence of
$\vartheta^{(n)}$ along any subsequence for which
$\Phi^{(n)}\to 0$ almost surely.

Fix such a subsequence $(n_m)$ and define the event
\[
\Omega^{(0)} := \bigcap_{j=1}^3 \Omega^{(j)},
\]

where
\[
\begin{aligned}
\Omega^{(1)}
&:= \left\{\Phi^{(n_m)} \to 0\right\},\\
\Omega^{(2)}
&:= \left\{
\max_{1\le i\le n}
\|\nabla \ell(\globpar^*;\obs{i},\cdot)\|
\le n^{1/p_2}
\ \text{a.b.f.o.}
\right\},\\
\Omega^{(3)}
&:= \left\{
\max_{1\le i\le n}
\|\nabla^{\otimes 2} \ell(\cdot;\obs{i},\cdot)\|_{\infty}
\le n^{1/p_3}
\ \text{a.b.f.o.}
\right\}.
\end{aligned}
\]
By the assumed moment conditions and Lemma~2 of
\citet{negrea2022statistical} applied to the power functions
$t\mapsto t^{p_2}$ and $t\mapsto t^{p_3}$, the event
$\Omega^{(0)}$ has probability one.
We will show that on $\Omega^{(0)}$, all remainder terms appearing in the generator
expansions are negligible, and the convergence of the discrete-time
Markov generators to their continuous-time limit follows.

\section{Proof of Main Theorem}

\subsection{Overview}

Our proof follows the spirit of \citet{negrea2022statistical}. 
We establish weak convergence
of the processes in the Skorokhod topology in probability by first proving almost sure
convergence along subsequences. This is achieved by showing that the difference between
the approximate generator and the limiting generator, evaluated on smooth test functions
with compact support, vanishes uniformly. Using Lemma~\ref{lem:subsequence_convergence}, this
then yields weak convergence in the Skorokhod topology in probability.
As in \citet{negrea2022statistical}, we divide the proof into two parts for technical reasons.

In Part~1, we consider arguments that are sufficiently far from the support of the test
function.
The main idea is to control the probability that the global-parameter process
jumps back into the support of the test function.
A key difference with \citet{negrea2022statistical} is that we must impose assumptions on the
joint likelihood uniformly on latent variable values to ensure that, even in the presence of additional uncertainty induced by the
latent-variable updates, the probability that the global parameter jumps back into the
support can still be controlled at a comparable scale.

In Part~2, we analyze arguments that lie in or are close to the support of the
test function.
We perform a Taylor expansion of the joint approximate generator. In \cref{subseq:R1}, as a result of
\cref{eq:l_marginalize}, the drift term converges in the same way as in
\citet{negrea2022statistical}. In \cref{subseq:R2}, for the gradient component of the diffusion term, a similar
approach applies, but the resulting limit now explicitly incorporates additional variability arising
from sampling the latent variables. In \cref{subseq:R3}, we introduce a new term that bridges the infinitesimal operator
associated with Gibbs updates and the generator of a Poisson jump process. The most
technically challenging new aspect is in \cref{subseq:R1,subseq:R5}, showing that all cross terms involving both the
global parameters and the latent variables vanish in the limit. This vanishing further
implies that, in the asymptotic regime, the contribution of any single observation becomes
asymptotically independent of the global-parameter dynamics, which in turn allows the
joint limiting distribution to factorize into independent components.

\subsection{Notation useful for the proof}

We first introduce notation for the increments of the localized algorithm.
Throughout, we condition on
$\vartheta^{(n)}_0=\vartheta$ and $\zeta^{(n)}_0=\zeta$,
and write $\zeta^{(n)}_1=\tilde\zeta$ for the updated latent variable.

Define the following components of the one-step increment:
\[
\Delta^{(n)}_{\xi}
:=
w^{(n)}
\sqrt{
h^{(n)} (\beta^{(n)})^{-1} \Gamma
}\,\xi_1 ,
\]
\[
\Delta^{(n)}_{\pi_0}
:=
\frac{h^{(n)} w^{(n)} \Gamma}{2n}
\nabla \log \pi_0\!\left(
\hat\theta^{(n)} + (w^{(n)})^{-1}\vartheta
\right),
\]
\[
\Delta^{(n)}_{\ell}
:=
\frac{h^{(n)} w^{(n)} \Gamma}{2 b^{(n)}}
\sum_{j=1}^{b^{(n)}}
\nabla_\theta
\ell\!\left(
\hat\theta^{(n)} + (w^{(n)})^{-1}\vartheta ;
X_{I^{(n)}_1(j)},
\tilde \zeta_{I^{(n)}_1(j)}
\right),
\]
and set
\[
\Delta^{(n)}
=
\Delta^{(n)}_{\xi}
+
\Delta^{(n)}_{\pi_0}
+
\Delta^{(n)}_{\ell}.
\]

The latent variables are updated according to
\[
\tilde \zeta_i
\sim
p\!\left(
z \mid X_i,
\hat\theta^{(n)} + (w^{(n)})^{-1}\vartheta
\right),
\qquad i=1,\dots,n ,
\]
independently conditional on the observations and the current parameter value.

We next define a sequence of generator-like operators acting on test functions
$f$ by
\[
A^{(n)} f(\vartheta,\zeta_1)
:=
\alpha^{(n)}
\Big(
\mathbb E\big[
f(\vartheta+\Delta^{(n)}, \tilde \zeta_1)
\big]
-
f(\vartheta,\zeta_1)
\Big),
\]
where the expectation is taken over all algorithmic randomness, including
minibatch sampling, Gaussian noise, and Gibbs updates, conditional on the
observations.

For sufficiently smooth test functions $f$, the generator $\gen $ of the
limiting Lévy process is given by
\[
\begin{aligned}
(\gen  f)(\vartheta,\zeta_1)
&=
- \langle B \vartheta, \nabla_\vartheta f(\vartheta,\zeta_1) \rangle
+ \frac{1}{2}
A : \nabla_\vartheta^{\otimes 2} f(\vartheta,\zeta_1) \\
&\quad
+ \lambda
\left(
\int f(\vartheta,z)\,
p(z \mid X_1,\theta^\star)\,dz
-
f(\vartheta,\zeta_1)
\right),
\end{aligned}
\]
with
\[
\begin{aligned}
B
&=
c_h \Gamma J_\star \mathbf{1}\{ \mathfrak{a} = \mathfrak{h} \},\\
A
&=
\frac{c_h}{c_\beta} \Gamma \mathbf{1}\{ \mathfrak{h}+\mathfrak{b}\leq \mathfrak{t} \}
+
\frac{c_h^2}{4c_b}
\Gamma \tilde{I_\star} \Gamma^\top
\mathbf{1}\{ \mathfrak{t} \leq \mathfrak{b}+\mathfrak{h} \}\\
\lambda& = c_b.
\end{aligned}
\]

Consider realization of $X^{(n)} \in \Omega^{(n)}$, we want to show that for all $f\in C^{\infty}_{c}(\mathbb{R}^{d+s})$ and any $\zeta_1$,
\[
\lim_{m \to \infty}\sup_{\limglob \in \mathbb{R}^d}\|\op f(\limglob,\limlat_1) - \gen f(\limglob,\limlat_1)\| = 0.
\]
We will show this in two parts.
To begin, note that, for any test function $f$ with compact support, there exists $K_0$ such that $f(\globpar)=0$ for all $\globpar \in K_0^c$.
First we will identify a extension set $K_1 \subset K_0$ 
such that 
\[
\lim_{m \to \infty}\sup_{\limglob \in K^{c}_{1}}\|\op f(\limglob,\limlat_1) - \gen f(\limglob,\limlat_1)\| = 0.
\]
Second, we will show that 
\[
\lim_{m \to \infty}\sup_{\limglob \in K_{1}}\|\op f(\limglob,\limlat_1) - \gen f(\limglob,\limlat_1)\| = 0.
\]
\subsection{Part 1.}
\label{sec:part1}

For all $\vartheta \in K^c_0$, we have
$f(\vartheta)=0$,
$\nabla_{\vartheta} f(\vartheta)=0$,
and
$\nabla_{\vartheta }^{\otimes 2} f(\vartheta)=0$.
Therefore, for any $K_1 \supseteq K_0$,
\[
\begin{aligned}
\sup_{\vartheta \in K^{c}_{1}}
\|\op f(\vartheta,\limlat_1) - \gen f(\vartheta,\limlat_1)\| 
&=
\timescalparn\sup_{\vartheta \in K^{c}_{1}}
\expect[f(\vartheta + \globincre,\newlat_1)] \\
&\le
\timescalparn \|f\|_{\infty}
\sup_{\vartheta \in K^{c}_{1}}
\mathbb{P}[\vartheta + \globincre \in K_0].
\end{aligned}
\]

Let
\[
K_1 \defas \{\vartheta : \|\vartheta\| \le 2R_0 +2c_0\},
\qquad
R_0 \defas \sup_{\vartheta \in K_0}\|\vartheta\|,
\]
where
\[
c_0 \defas
\frac{\conststep}{2}(3+\|\precon\nabla \log \prior(\globpar^*)\|)
+\sqrt{\conststep /c_{\beta}\Gamma}.
\]

Then, for $\vartheta \in K_1^c$, under the assumption that
$\precon\nabla\log\prior$ is $L_0$-Lipschitz,
\[
\begin{aligned}
\|\globincreprior\|
&=
\frac{\stepsizen \spascaln \precon}{2\samplesize}
\|\nabla\log \prior(\mleglobn+(\spascaln)^{-1}\vartheta)\| \\
&\le
\frac{\conststep \samplesize^{\powerspa -\powerstep-1}}{2}
\Bigl(
\|\precon\nabla\log\prior(\globpar^*)\|
+ L_0 \|\mleglobn -\globpar^*\|
+ \frac{L_0 \|\vartheta\|}{n^{\powerspa}}
\Bigr).
\end{aligned}
\]

Similarly,
\[
\begin{aligned}
\|\globincrell\|
&=
\frac{\stepsizen \spascaln \|\precon\|}{2 \batchsizen}
\Bigl\|
\sum_{\obsidx=1}^{\batchsizen}
\nabla \ell(\mleglobn+(\spascaln)^{-1}\vartheta;
\obs{\randidx_{1}^{(\samplesize)}(\obsidx)},
\newlat_{\randidx_{1}^{(\samplesize)}(\obsidx)})
\Bigr\| \\
&\le
\frac{\conststep n^{\powerspa-\powerstep}\|\precon\|}{2 \batchsizen}
\sum_{\obsidx\in[\subseq]}
\|\nabla \ell(\globpar^*;
\obs{\randidx_{1}^{(\subseq)}(\obsidx)},\newlat_{\randidx_{1}^{(\subseq)}(\obsidx)})\| \\
&\quad
+ L(\obs{\randidx_{1}^{(\subseq)}(\obsidx)},\newlat_{\randidx_{1}^{(\subseq)}(\obsidx)})
\|\mleglobn-\globpar^*\| \\
&\quad
+ L(\obs{\randidx_{1}^{(\subseq)}(\obsidx)},\newlat_{\randidx_{1}^{(\subseq)}(\obsidx)})
\frac{\|\vartheta\|}{\subseq^{\powerspa}} .
\end{aligned}
\]

We define the (random) Lipschitz constants
\[
\begin{aligned}
L(\obs{i})
&:=\|\nabla^{\otimes 2} \ell(\cdot;\obs{i},\cdot)\|_{\infty},\\
L_{*}(\ob^{(\subseq)})
&:=\max_{i \in [\subseq]}\|\nabla \ell(\globpar^*;\obs{i},\cdot)\|,\\
L(\ob^{(\subseq)},\newlat^{(\subseq)})
&:= \max_{\obsidx\in[\subseq]}L(\obs{i}).
\end{aligned}
\]

Since $\ob^{(\subseq)} \in \Omega^{(0)}$, we have that $\Phi^{(\subseq)} \to 0$, $\max_{i \in [\subseq]}\|\nabla \ell(\globpar^*;\obs{i},\cdot)\|_{\infty}
\le \subseq^{1/p_2}$, and $\max_{i \in [\subseq]}
\|\nabla^{\otimes 2} \ell(\cdot;\obs{i},\cdot)\|_{\infty}
\le \subseq^{1/p_3}$. Thus if $m$ is large enough that all of the following hold:
\begin{align}
\sup_{m'\ge m}\Phi^{(\subseq)} &\le \min(1,L_{0}^{-1}),\\
 1 &\ge \sup_{m'\ge m} \frac{L_{*}(\ob^{(n_{m'})})}{n_{m'}^{1/p_{2}}}\\
\subseq &\ge \max \left((2\conststep \|\precon\|)^{1/(1/p_{3}-\powerstep)},(2\conststep L_0 \|\precon\|)^{\frac{1}{\powerstep+1-\powertime-\powerspa}}\right)\\
1 &\ge \sup_{m'\ge m} \frac{L(\ob^{(n_{m'})})}{n_{m'}^{1/p_{3}}}.
\end{align}

Then, using that $0 < \powerspa < 1$,
\[
\|\globincreprior\|
\le
\frac{c_h\|\precon\|}{2}\bigl(\|\nabla\log\prior(\globpar^*)\|+1\bigr)
+\frac{1}{4}\|\vartheta\|,
\qquad
\|\globincrell\|
\le
c_h\|\precon\|+\frac{1}{4}\|\vartheta\|.
\]

Therefore, for $\vartheta \in K_1^c$,
\[
\|\vartheta\|
-\|\globincreprior\|
-\|\globincrell\|
-R_0
\ge
\frac{1}{2}\|\vartheta\|
-\frac{c_h\|\precon\|}{2}(3+\|\nabla\log\prior(\globpar^*)\|)
-R_0
\ge
\sqrt{c_h\|\precon\|/c_\beta}.
\]

Consequently,
\[
\begin{aligned}
\lim_{m \to \infty}
\sup_{\vartheta \in K^{c}_{1}}
\|\op f(\vartheta,\limlat_1) - \gen f(\vartheta,\limlat_1)\| 
&\le
\lim_{m \to \infty}
\alpha^{\subseq}
\|f\|_{\infty}
\mathbb{P}\!\left(
\|\xi_1\|\ge
\subseq^{\powerstep/2+\powerit/2-\powerspa}
\right)
=0.
\end{aligned}
\]

\subsection{Part 2.}
\label{sec:part2}

We take a partial second-order Taylor expansion of the test function $f$
with respect to the global variable $\vartheta$:
\[
\begin{aligned}
\op f(\vartheta,\limlat_1)
&=
\timescalparn
\bigl(
\expect[f(\vartheta + \globincre,\newlat_1)]
-
f(\vartheta,\limlat_1)
\bigr) \\
&=
\timescalparn
\Bigl(
\expect[f(\vartheta + \globincre,\newlat_1)-f(\vartheta,\newlat_1)]
+
\expect[f(\vartheta,\newlat_1)-f(\vartheta,\limlat_1)]
\Bigr) \\
&=
\subseq^{\powertime}
\expect\innerprod{\nabla_\vartheta f(\vartheta ,\newlat_1)}{\globincre}
+
\subseq^{\powertime}
\expect\innerprod{\frac{1}{2}\nabla_{\vartheta}^{\otimes 2}f(\vartheta ,\newlat_1)\globincre}{\globincre} \\
&\quad
+
\subseq^{\powertime}
\expect\!\left[
\frac{1}{6}
\nabla_{\vartheta}^{\otimes 3}
f(\vartheta+S\globincre,\newlat_1)
(\globincre,\globincre,\globincre)
\right] \\
&\quad
+
\expect[f(\vartheta,\newlat_1)-f(\vartheta,\limlat_1)] .
\end{aligned}
\]
Rearranging terms yields
\[
\begin{aligned}
\op f(\vartheta,\limlat_1)
&=
\subseq^{\powertime}
\expect\innerprod{\nabla_\vartheta f(\vartheta ,\limlat_1)}{\globincre}
+
\subseq^{\powertime}
\expect\innerprod{\frac{1}{2}\nabla_{\vartheta}^{\otimes 2}f(\vartheta ,\limlat_1)\globincre}{\globincre} \\
&\quad
+
\subseq^{\powertime}
\expect\!\left[
\frac{1}{6}
\nabla_{\vartheta}^{\otimes 3}
f(\vartheta+S\globincre,\newlat_1)
(\globincre,\globincre,\globincre)
\right] \\
&\quad
+
\expect[f(\vartheta,\newlat_1)-f(\vartheta,\limlat_1)] \\
&\quad
+
\subseq^{\powertime}
\expect\innerprod{
\nabla_\vartheta f(\vartheta ,\newlat_1)
-
\nabla_\vartheta f(\vartheta ,\limlat_1)
}{\globincre} \\
&\quad
+
\subseq^{\powertime}
\expect\innerprod{
\frac{1}{2}\nabla_{\vartheta}^{\otimes 2}f(\vartheta ,\newlat_1)\globincre
-
\frac{1}{2}\nabla_{\vartheta}^{\otimes 2}f(\vartheta ,\limlat_1)\globincre
}{\globincre}
\end{aligned}
\]
for some $S \in [0,1]$.
Therefore,
\begin{align}
\|\op f - \gen f\|
\le\;
&\underbrace{
\Bigl\|
\subseq^{\powertime}
\expect\innerprod{\nabla_\vartheta f(\vartheta ,\limlat_1)}{\globincre}
+
\innerprod{\frac{1}{2}B \vartheta}{ \nabla_\vartheta f(\vartheta ,\limlat_1)}
\Bigr\|
}_{R_1} \\
&+
\underbrace{
\Bigl\|
\subseq^{\powertime}
\expect\innerprod{
\frac{1}{2}\nabla_{\vartheta}^{\otimes 2}f(\vartheta ,\limlat_1)\globincre
}{\globincre}
-
\frac{1}{2} A:\nabla_{\vartheta}^{\otimes 2}f(\vartheta ,\limlat_1)
\Bigr\|
}_{R_2} \\
&+
\underbrace{
\Bigl\|
\lambda\Bigl(
\int f(\vartheta,y)p(y|\obs{1},\globpar^*)\,dy
-
f(\vartheta,\limlat_1)
\Bigr)
-
\subseq^{\powertime}
\expect[f(\vartheta,\newlat_1)-f(\vartheta,\limlat_1)]
\Bigr\|
}_{R_3} \\
&+
\underbrace{\Bigl\|
\subseq^{\powertime}
\expect\innerprod{
\nabla_\vartheta f(\vartheta ,\newlat_1)
-
\nabla_\vartheta f(\vartheta ,\limlat_1)
}{\globincre}\Bigr\|
}_{R_4} \\
&+
\underbrace{\Bigl\|
\subseq^{\powertime}
\expect\innerprod{
\frac{1}{2}\nabla_{\vartheta}^{\otimes 2}f(\vartheta ,\newlat_1)\globincre
-
\frac{1}{2}\nabla_{\vartheta}^{\otimes 2}f(\vartheta ,\limlat_1)\globincre
}{\globincre}\Bigr\|
}_{R_5} \\
&+
\underbrace{\Bigl\|
\subseq^{\powertime}
\expect\!\left[
\frac{1}{6}
\nabla_{\vartheta}^{\otimes 3}
f(\vartheta+S\globincre,\newlat_1)
(\globincre,\globincre,\globincre)
\right]\Bigr\|
}_{R_6}.
\end{align}

We denote the six terms above by $R_1,\dots,R_6$ and will show that each of them
vanishes as $n \to \infty$.

\subsection{$R_1$ (drift term)}
\label{subseq:R1}
We have
\[
\subseq^{\powertime}\,\mathbb{E}[\globincre]
=
\subseq^{\powertime}\,\mathbb{E}\bigl[\globincrenoise+\globincreprior+\globincrell\bigr].
\]

\paragraph{Noise term.}
\[
\subseq^{\powertime}\,\mathbb{E}[\globincrenoise]
=
\subseq^{\powertime}\,\mathbb{E}\!\left[\spascaln \sqrt{\stepsizen (\invtemp)^{-1}\precon}\,\xi_1\right]
=
\subseq^{\powertime}\,\spascaln \sqrt{\stepsizen (\invtemp)^{-1}\precon}\,\mathbb{E}[\xi_1]
=
0.
\]

\paragraph{Prior term.}
\begin{align}
\subseq^{\powertime}\,\mathbb{E}[\globincreprior]
&=
\subseq^{\powertime}\,
\mathbb{E}\!\left[
\frac{\stepsizen \spascaln \precon}{2\samplesize}\,
\nabla\log \prior\!\bigl(\mleglobn+(\spascaln)^{-1}\limglob\bigr)
\right] \\
&=
\frac{\conststep}{2}\,
\subseq^{\powertime-\powerstep+\powerspa-1}\,
\precon\,
\nabla\log \prior\!\bigl(\mleglobn+(\spascaln)^{-1}\limglob\bigr)\\
&=
\frac{\conststep \subseq^{\powertime-\powerstep+\powerspa-1}\|\precon\|}{2}\,
\Bigg(\nabla\log \prior\!\bigl(\globpar^*)+L_0(\Phi^{(\subseq)} + \frac{2R_0+2c_0}{\subseq^\powerspa})\Bigg).
\end{align}
Which vanishes uniformly on $K_1$, as $\powertime-\powerstep+\powerspa-1<0$

\paragraph{Likelihood term.}
\begin{align}
\subseq^{\powertime}\expect\globincrell 
& =  n^{\powertime}\expect\frac{\stepsizen \spascaln\precon}{2 \batchsizen}\sum_{\obsidx=1}^{\batchsizen}\nabla \ell(\mleglobn+(\spascaln)^{-1}\limglob;\obs{\randidx_{1}^{(\samplesize)}(\obsidx)},\newlat_{\randidx_{1}^{(\samplesize)}(\obsidx)})\\
& = \frac{\conststep n^{\powertime-\powerstep+\powerspa-1}\precon}{2 }\expect_{\newlat}\sum_{\obsidx=1}^{\subseq}\nabla \ell(\mleglobn+\subseq^{-\powerspa}\limglob;\obs{\obsidx},\newlat_{\obsidx})\\
& = \frac{\conststep \subseq^{\powertime-\powerstep+\powerspa-1}\precon}{2 }\sum_{\obsidx=1}^{\subseq}\nabla \ell(\mleglobn+n^{-\powerspa}\limglob;\obs{\obsidx})\\
& = \frac{\conststep \subseq^{\powertime-\powerstep+\powerspa-1}\precon}{2 }\sum_{\obsidx=1}^{\subseq}[\nabla \ell(\mleglobn;\obs{\obsidx})+n^{-\powerspa}\nabla^{2}\ell(\mleglobn;\obs{\obsidx})\limglob]\\
& = \frac{\conststep \subseq^{\powertime-\powerstep}\precon}{2}\frac{\limglob}{\subseq}\sum_{\obsidx=1}^{\subseq}\int_{0}^{1}\nabla^{\otimes 2}\ell(\mleglobn+\frac{s}{\subseq^{\powerspa}\limglob};\obs{\obsidx})ds
\end{align}
where we used that $\ell(\globpar;\obs{\obsidx})$ denotes the (marginal) log-likelihood with $\latvar$ marginalized, and
\[
\mathbb{E}_{\newlat}[g(\newlat_{\obsidx})]
=
\int g(\newlat_{\obsidx})\,p(\newlat_{\obsidx}\mid \obs{\obsidx},\mleglobn+\subseq^{-\powerspa}\limglob)\,d\newlat_{\obsidx}.
\]

Indeed,
\begin{align}
\label{eq:l_marginalize}
\mathbb{E}_{\newlat}\!\left[\nabla \ell(\mleglobn+\subseq^{-\powerspa}\limglob;\obs{\obsidx},\newlat_{\obsidx})\right]
&=
\int \nabla \log p(\obs{\obsidx},\newlat_{\obsidx}\mid \mleglobn+\subseq^{-\powerspa}\limglob)\,
p(\newlat_{\obsidx}\mid \obs{\obsidx},\mleglobn+\subseq^{-\powerspa}\limglob)\,d\newlat_{\obsidx} \\
&=
\frac{\nabla p(\obs{\obsidx}\mid \mleglobn+\subseq^{-\powerspa}\limglob)}
     {p(\obs{\obsidx}\mid \mleglobn+\subseq^{-\powerspa}\limglob)} \\
&=
\nabla \ell(\mleglobn+\subseq^{-\powerspa}\limglob;\obs{\obsidx}).
\end{align}

Moreover, by the definition of $\mleglobn$,
\[
\sum_{\obsidx=1}^{\subseq}\nabla \ell(\mleglobn;\obs{\obsidx}) = 0.
\]
Hence, when $\powertime +\powerstep <1$, $\subseq^{\powertime}\expect\globincrell $ vanishes and the drift term will be inactive in the limit.

When $\powertime =\powerstep $, and $\subseq$ large enough that $r_{J,\subseq}\geq R_0+c_0$,
\begin{align}
R_1 = \Bigl\|\subseq^{\powertime}
\expect\innerprod{\nabla_\vartheta f(\vartheta ,\limlat_1)}{\globincre}
+
\innerprod{\frac{c_h\precon J_\star}{2} \vartheta}{ \nabla_\vartheta f(\vartheta ,\limlat_1)}\Bigr\|\leq c_h \|f\|_{\infty}\|\precon\|(R_0+c_0)\Phi^{(\subseq)},
\end{align}
which vanishes uniformly on $K_1$

\subsection{$R_2$ (diffusion term)}
\label{subseq:R2}
We write
\[
\subseq^{\powertime}\,\mathbb{E}\!\left[(\globincre)^{\otimes 2}\right]
=
\subseq^{\powertime}\,
\mathbb{E}\!\left[(\globincrenoise+\globincreprior+\globincrell)^{\otimes 2}\right].
\]
Given the independence among all three terms, the cross term will vanish. So the potentially non-vanishing contributions come from
$\subseq^{\powertime}\mathbb{E}\!\left[(\globincrenoise)^{\otimes 2}\right]$
and
$\subseq^{\powertime}\mathbb{E}\!\left[(\globincrell)^{\otimes 2}\right]$.

To obtain a non-trivial limit, we only considering $\powertime =\powerstep$.

\paragraph{Noise variance.}
\[
\subseq^{\powertime}\,\mathbb{E}\!\left[(\globincrenoise)^{\otimes 2}\right]
=
\frac{\conststep}{c_{\beta}}\,
\subseq^{\powertime+2\powerspa-\powerstep-\powerit}\,\precon
=
\frac{\conststep}{c_{\beta}}\,
\subseq^{2\powerspa-\powerstep}\,\precon.
\]

Thus when $\powerspa <\powerit/2$, the corresponding diffusion term is inactive in the limit. When $\powerspa =\powerit/2$,
\[
\|\subseq^{\powertime}\,\mathbb{E}\!\left[(\globincrenoise)^{\otimes 2}\right]- \frac{c_h\precon}{c_\beta}\|= 0 .
\]
\paragraph{Stochastic gradient variance.}
\begin{align}
n^\powertime \expect[\globincrell]^2 & =  n^\powertime \expect\left[\frac{\stepsizen \spascaln}{2 \batchsizen}\sum_{\obsidx=1}^{\batchsizen}\nabla \ell(\mleglobn+(\spascaln)^{-1}\limglob;\obs{\randidx_{1}^{(\samplesize)}(\obsidx)},\newlat_{\randidx_{1}^{(\samplesize)}(\obsidx)})\right]^2\\
& =\frac{\conststep^2}{4\constbatch}n^{\powertime-2\powerstep+2\powerspa-2\powerbatch}\precon\expect\left[\sum_{\obsidx=1}^{\batchsizen}\nabla \ell(\mleglobn+(\spascaln)^{-1}\limglob;\obs{\randidx_{1}^{(\samplesize)}(\obsidx)},\newlat_{\randidx_{1}^{(\samplesize)}(\obsidx)})\right]^2 \precon'\\
& = \frac{\conststep^2}{4\constbatch}n^{\powertime-2\powerstep+2\powerspa-\powerbatch}\precon\left[\frac{1}{\subseq}\expect_{\newlat}\sum_{\obsidx=1}^{\subseq}\nabla \ell(\mleglobn+(\spascaln)^{-1}\limglob;\obs{\obsidx},\newlat_{\obsidx})^2\right]\precon' \\
&+ \frac{\conststep^2}{4\constbatch^2}n^{\powertime-2\powerstep+2\powerspa-2\powerbatch}\precon\Bigg[\expect\sum_{\obsidx=1}^{\batchsizen}\sum_{\obsidx '=1, \obsidx '\ne \obsidx}^{\batchsizen}\nabla \ell(\mleglobn+(\spascaln)^{-1}\limglob;\obs{\randidx_{1}^{(\samplesize)}(\obsidx)},\newlat_{\randidx_{1}^{(\samplesize)}(\obsidx)})\\
&\phantom{\frac{\conststep^2}{4\constbatch^2}n^{\powertime-2\powerstep+2\powerspa-2\powerbatch}\precon\Bigg[\expect\sum_{\obsidx=1}^{\batchsizen}\sum_{\obsidx '=1, \obsidx '\ne \obsidx}^{\batchsizen}}\otimes\nabla \ell(\mleglobn+(\spascaln)^{-1}\limglob;\obs{\randidx_{1}^{(\samplesize)}(\obsidx ')},\newlat_{\randidx_{1}^{(\samplesize)}(\obsidx ')})\Bigg]\precon'\\
& = \frac{\conststep^2}{4\constbatch}\subseq^{\powertime-2\powerstep+2\powerspa-\powerbatch}\precon\Bigg[\frac{1}{\subseq}\expect_{\newlat}\sum_{\obsidx=1}^{\subseq}\nabla \ell(\mleglobn+(\spascaln)^{-1}\limglob;\obs{\obsidx},\newlat_{\obsidx})^2\Bigg]\precon' \\
&+ \frac{\conststep^2}{4\constbatch^2}\subseq^{\powertime-2\powerstep+2\powerspa-2\powerbatch}\frac{\batchsizen(\batchsizen-1)}{\subseq^2}\precon\Bigg[\sum_{\obsidx=1}^{\subseq}\sum_{\obsidx '=1}^{\subseq}\nabla \ell(\mleglobn+(\spascaln)^{-1}\limglob;\obs{\randidx_{1}^{(\samplesize)}(\obsidx)})\\&\phantom{\frac{\conststep^2}{4\constbatch^2}\subseq^{\powertime-2\powerstep+2\powerspa-2\powerbatch}\frac{\batchsizen(\batchsizen-1)}{\subseq^2}\precon\sum_{\obsidx=1}^{\subseq}\sum_{\obsidx '=1}^{\subseq}}\otimes\nabla \ell(\mleglobn+(\spascaln)^{-1}\limglob;\obs{\randidx_{1}^{(\samplesize)}(\obsidx ')})\Bigg]\precon'\\
& = \frac{\conststep^2}{4\constbatch}\subseq^{\powertime-2\powerstep+2\powerspa-\powerbatch}\precon\Bigg[\frac{1}{\subseq}\expect_{\newlat}\sum_{\obsidx=1}^{\subseq}\nabla \ell(\mleglobn+(\spascaln)^{-1}\limglob;\obs{\obsidx},\newlat_{\obsidx})^2\Bigg]\precon'\\
&+ \frac{\conststep^2}{4\constbatch^2}\subseq^{\powertime-2\powerstep+2\powerspa-2\powerbatch}\batchsizen(\batchsizen-1)\precon\Bigg(\frac{1}{\subseq}\sum_{\obsidx=1}^{\subseq}\int_{0}^{1}\nabla^{\otimes 2}\ell(\mleglobn+\frac{s}{\subseq^{\powerspa}}\limglob;\obs{\obsidx})ds\frac{1}{\subseq^{\powerspa}}\limglob\Bigg)^{\otimes 2}\precon'.
\end{align}

Note that 
\begin{align}
&\bigg\|\frac{\conststep^2}{4\constbatch^2}\subseq^{\powertime-2\powerstep+2\powerspa-2\powerbatch}\batchsizen(\batchsizen-1)\precon\Bigg(\frac{1}{\subseq}\sum_{\obsidx=1}^{\subseq}\int_{0}^{1}\nabla^{\otimes 2}\ell(\mleglobn+\frac{s}{\subseq^{\powerspa}}\limglob;\obs{\obsidx})ds\frac{1}{\subseq^{\powerspa}}\limglob\Bigg)^{\otimes 2}\precon f'\bigg\|\\
&\quad \leq \sqrt{d}\conststep^2 \subseq^{\powertime-2\powerstep+2\powerspa}\|\precon\|^2\|\nabla^{\otimes 2}f\|_{\infty} \frac{(2R_0+2c_0)^2}{\subseq^{2\powerspa}}(J_\star+\Phi^{(\subseq)})^2
\end{align}

Since $\powertime =\powerstep$ and $\powerstep >0$, this term vanishes uniformly on $K_1$.

Thus when $\powerspa <(\powerstep+\powerbatch)/2$, the corresponding diffusion term is inactive in the limit. When $\powerspa <(\powerstep+\powerbatch)/2$ and $\subseq$ large enough that $r_{I,\subseq}\geq R_0+c_0$,
\[
\|n^\powertime \expect[\globincrell]^2-\frac{c_h^2}{4c_b}\precon \tilde I_\star \precon'\| \leq \frac{c_h^2}{4c_b}\|\precon\|^2\Phi^{(\subseq)}
\]
vanishes uniformly on $K_1$.
Hence, $R_2$ 
vanishes uniformly on $K_1$.

\subsection{$R_3$ (jump term)}
\label{subseq:R3}
We consider
\begin{align}
\subseq^{\powertime}\,\mathbb{E}\!\left[f(\limglob,\newlat_1)-f(\limglob,\limlat_1)\right]
&=
\constbatch\,\subseq^{\powertime+\powerbatch-1}
\left[
\int f(\limglob,y)\,p(y\mid \obs{1},\mleglobn+\subseq^{-\powerspa}\limglob)\,d\zeta
-
f(\limglob,\limlat_1)
\right] \\
&=
\constbatch\,\subseq^{\powertime+\powerbatch-1}
\left[
\int f(\limglob,y)\,p(y\mid \obs{1},\globpar^*)\,d\zeta
-
f(\limglob,\limlat_1)
\right] \\
&\quad
+\constbatch\,\subseq^{\powertime+\powerbatch-1}
\left[
\int f(\limglob,y)\,
\bigl(p(y\mid \obs{1},\mleglobn+\subseq^{-\powerspa}\limglob)-p(y\mid \obs{1},\globpar^*)\bigr)\,dy
\right].
\end{align}

 Since $\|\mleglobn+\subseq^{-\powerspa}\limglob-\globpar^*\| \leq \|\mleglobn-\globpar^*\|+\subseq^{-\powerspa}\|\limglob\|$ vanishes uniformly on $K_1$, under \cref{asp6}, so do $\|p(y\mid \obs{1},\mleglobn+\subseq^{-\powerspa}\limglob)-p(y\mid \obs{1},\globpar^*)\|$.

 Thus when $\powerbatch+\powerstep<1$, this term is inactive in the limit. 
 When $\powerbatch+\powerstep=1$,
\[
R_3 = \|
\lambda\Bigl(
\int f(\vartheta,y)p(y|\obs{1},\globpar^*)\,dy
-
f(\vartheta,\limlat_1)
\Bigr)
-
\subseq^{\powertime}
\expect[f(\vartheta,\newlat_1)-f(\vartheta,\limlat_1)]\|
\]
vanishes uniformly on $K_1$ with $\lambda = c_b.$

\subsection{$R_4$ (Gradient mismatch term)}
\label{subseq:R4}
Recall that
\(
\nabla_\limglob f(\limglob,\newlat_1)
-
\nabla_\limglob f(\limglob,\limlat_1)
\)
is non-zero only when the latent variable $\zeta_1$ is updated, i.e.,
when
\[
A_1 \defas \{1 \in I_1^{(\subseq)}\}
\]
occurs. Therefore, we analyze
\[
\label{eq:gradientmm}
\subseq^{\powertime}\,
\mathbb{E}\!\left[
\bigl\langle
\nabla_\limglob f(\limglob,\newlat_1)
-
\nabla_\limglob f(\limglob,\limlat_1),
\globincre
\bigr\rangle
\right]
\]
by conditioning on $A_1$.

By the law of total expectation,
\begin{align}
&\subseq^{\powertime}\,
\mathbb{E}\!\left[
\bigl\langle
\nabla_\limglob f(\limglob,\newlat_1)
-
\nabla_\limglob f(\limglob,\limlat_1),
\globincre
\bigr\rangle
\right]\notag\\
&=
\subseq^{\powertime}\,
\mathbb{P}(A_1)\,
\mathbb{E}\!\left[
\bigl\langle
\nabla_\limglob f(\limglob,\newlat_1)
-
\nabla_\limglob f(\limglob,\limlat_1),
\globincre
\bigr\rangle
\;\middle|\;
A_1
\right].
\label{eq:R4_conditional}
\end{align}
Since the mini-batch is sampled with replacement,
\[
\mathbb{P}(A_1)
=
1-\Bigl(1-\frac{1}{\subseq}\Bigr)^{\batchsizen}.
\]

Note that $\globincrenoise$ and $\globincreprior$ is independent of $A_1$, thus $\nabla_\limglob f(\limglob,\newlat_1)
-
\nabla_\limglob f(\limglob,\limlat_1)$ brings in a factor that is bounded by $\subseq^{\powerbatch-1}\|\nabla f\|_{\infty}$, keeping the effects from Gaussian noise and prior still inactive (As discussed in \cref{subseq:R1}) in the limit, thus we only consider $\globincrell $.
Under $A_1$, the conditional expectation of the increment satisfies
\begin{align}
\mathbb{E}[\globincrell \mid A_1]
&=
\frac{c_h\,\subseq^{-\powerstep+\powerspa}}{2}
\Biggl[
\frac{1}{\subseq\mathbb{P}(A_1)}
\,\precon\nabla\ell\!\bigl(
\mleglobn+(\spascaln)^{-1}\limglob;\obs{1},\tilde\zeta_1
\bigr)
\notag\\
&\qquad\qquad\quad
+
\Bigl(
1-\frac{1}{\subseq\mathbb{P}(A_1)}
\Bigr)
\frac{1}{\subseq-1}
\sum_{i=2}^{\subseq}
\precon\nabla\ell\!\bigl(
\mleglobn+(\spascaln)^{-1}\limglob;\obs{i},\tilde\zeta_i
\bigr)
\Biggr].
\label{eq:R4_conditional_increment}
\end{align}
Substituting \eqref{eq:R4_conditional_increment} into
\eqref{eq:R4_conditional} yields an explicit decomposition of the
gradient mismatch term.

Assume $\powerbatch-1<0$, so that $\batchsizen/\subseq \to 0$. Then
\[
\mathbb{P}(A_1)
=
1-\Bigl(1-\frac{1}{\subseq}\Bigr)^{\batchsizen}
=
\frac{\batchsizen}{\subseq}
+ O\!\left(\frac{{\batchsizen}^2}{\subseq^2}\right).
\]
Moreover,
\[
\frac{1}{\subseq\mathbb{P}(A_1)}
=
\frac{1}{\batchsizen}
+ o\left(\frac{1}{\batchsizen}\right).
\]

Assuming $\powertime=\powerstep$, the dominant contribution becomes
\begin{align}
\label{eq:gradientmm}
&\subseq^{\powertime}\,
\mathbb{E}\!\left[
\bigl\langle
\nabla_\limglob f(\limglob,\newlat_1)
-
\nabla_\limglob f(\limglob,\limlat_1),
\globincre
\bigr\rangle
\right]\notag\\
&=
\frac{c_h\precon}{2}\,
\subseq^{\powerspa-1}
\mathbb{E}_{\tilde\zeta}\!\left[
\bigl\langle
\nabla_\limglob f(\limglob,\newlat_1)
-
\nabla_\limglob f(\limglob,\limlat_1),
\nabla\ell\!\bigl(
\mleglobn+(\spascaln)^{-1}\limglob;\obs{1},\tilde\zeta_1
\bigr)
\bigr\rangle
\right]\notag\\
&\quad
+
\frac{c_h\precon}{2}\,
\subseq^{\powerspa-1}
\Biggl\langle
\mathbb{E}_{\tilde\zeta}
\!\left[
\nabla_\limglob f(\limglob,\newlat_1)
-
\nabla_\limglob f(\limglob,\limlat_1)
\right],
\frac{\batchsizen-1}{\subseq-1}
\sum_{i=2}^{\subseq}
\nabla\ell\!\bigl(
\mleglobn+(\spascaln)^{-1}\limglob;\obs{i}
\bigr)
\Biggr\rangle.
\end{align}

Note that
\begin{align}
\frac{1}{\subseq}\sum_{i=2}^{\subseq}
\nabla\ell\!\bigl(
\mleglobn+(\spascaln)^{-1}\limglob;\obs{i}
\bigr)
&=
\frac{1}{\subseq}\sum_{i=1}^{\subseq}
\nabla\ell\!\bigl(
\mleglobn+(\spascaln)^{-1}\limglob;\obs{i}
\bigr)
-
\frac{1}{\subseq}
\nabla\ell\!\bigl(
\mleglobn+(\spascaln)^{-1}\limglob;\obs{1}
\bigr)\notag\\
&=
\subseq^{-\powerspa} (J_*\,\limglob
+ R_J)
-
\subseq^{-1}
\nabla\ell\!\bigl(
\mleglobn+(\spascaln)^{-1}\limglob;\obs{1}
\bigr),
\end{align}
where $R_J$ vanishes uniformly on $K_1$ as $\subseq\to\infty$.
So only consider the dominant terms,
\begin{align}
\eqref{eq:gradientmm} \lesssim &\subseq^{\powerspa-1}\|\nabla f\|_{\infty}
\mathbb{E}_{\tilde\zeta}\!
\|\nabla\ell\!\bigl(
\mleglobn+(\spascaln)^{-1}\limglob;\obs{1},\tilde\zeta_1
\bigr)\|
\notag\\
&+ \|\nabla f\|_{\infty}\left(\subseq^{b-1}J_\star \limglob + \subseq^{\powerbatch+\powerspa-2}\nabla\ell\!\bigl(
\mleglobn+(\spascaln)^{-1}\limglob;\obs{1}
\bigr)\right)
\end{align}
Since $\powerspa-1 < 0$ and $\powerbatch-1 < 0$, and for $m$ large enough, $\max_{1\le i\le \subseq}
\|\nabla \ell(\globpar^*;\obs{i},\cdot)\|
\le \subseq^{1/p_2}$, all terms in \eqref{eq:gradientmm} vanish, 
$R_4$ vanishes uniformly on $K_1$.

\subsection{$R_5$ (Hessian mismatch term)}
\label{subseq:R5}
Let
\[
A_1 := \{1\in I_1^{(\subseq)}\},
\qquad
\mathbb P(A_1)=1-\Bigl(1-\frac{1}{\subseq}\Bigr)^{\batchsizen}.
\]
Again we will focus on 
\begin{align}
&\subseq^{\powertime}\,
\mathbb{E}\!\left[
(\globincrell)^{\otimes 2} :
\bigl(\nabla_{\limglob\limglob} f(\limglob,\newlat_1)-\nabla_{\limglob\limglob} f(\limglob,\limlat_1)\bigr)
\right]\notag\\
&=
\mathbb P(A_1)\,\subseq^{\powertime}\,
\mathbb{E}\!\left[
(\globincrell)^{\otimes 2} :
\bigl(\nabla_{\limglob\limglob} f(\limglob,\newlat_1)-\nabla_{\limglob\limglob} f(\limglob,\limlat_1)\bigr)
\;\middle|\;A_1
\right].\label{eq:R5_cond_start}
\end{align}

Under the event $A_1=\{1\in I_1^{(\subseq)}\}$, the conditional second moment
of the likelihood increment admits the decomposition
\begin{align}
&\mathbb{E}\!\left[(\globincrell)^{\otimes 2}\mid A_1\right]\\
&=
\left(\frac{c_h\,\subseq^{-\powerstep+\powerspa}}{2}\right)^2 \precon\Bigg\{\frac{1}{\batchsizen}
\Bigg[
\frac{\nabla\ell\!\bigl(
\mleglobn+(\spascaln)^{-1}\limglob;\obs{1},\tilde\zeta_1
\bigr)^{\otimes 2}}{\subseq\,\mathbb P(A_1)}
\\
&+
\left(1-\frac{1}{\subseq\,\mathbb P(A_1)}\right)
\frac{1}{\subseq-1}\sum_{i=2}^{\subseq} \nabla\ell\!\bigl(
\mleglobn+(\spascaln)^{-1}\limglob;\obs{i},\tilde\zeta_i
\bigr)^{\otimes 2}
\Bigg] \\
&+
\frac{(\batchsizen-1)}{\batchsizen}
\Bigg[
\left(\frac{1}{\subseq\,\mathbb P(A_1)}\right)^2 \nabla\ell\!\bigl(
\mleglobn+(\spascaln)^{-1}\limglob;\obs{1},\tilde\zeta_1
\bigr)^{\otimes 2}
\\
&+
\frac{2}{\subseq\,\mathbb P(A_1)}\left(1-\frac{1}{\subseq\,\mathbb P(A_1)}\right) \nabla\ell\!\bigl(
\mleglobn+(\spascaln)^{-1}\limglob;\obs{1},\tilde\zeta_1
\bigr)\otimes\left(\frac{1}{\subseq-1}\sum_{i=2}^{\subseq} \nabla\ell\!\bigl(
\mleglobn+(\spascaln)^{-1}\limglob;\obs{i},\tilde\zeta_i
\bigr)\right)
\\
&+
\left(1-\frac{1}{\subseq\,\mathbb P(A_1)}\right)^2 \left(\frac{1}{\subseq-1}\sum_{i=2}^{\subseq} \nabla\ell\!\bigl(
\mleglobn+(\spascaln)^{-1}\limglob;\obs{i},\tilde\zeta_i
\bigr)\right)^{\otimes 2}\Bigg] \Bigg\}\precon'.
\label{eq:R5_conditional_second_moment}
\end{align}

Assuming $\powertime=\powerstep$, the dominant contribution of 
\begin{align}
&\subseq^{\powertime}\,
\mathbb{E}\!\left[
(\globincrell)^{\otimes 2} :
\bigl(\nabla_{\limglob\limglob} f(\limglob,\newlat_1)-\nabla_{\limglob\limglob} f(\limglob,\limlat_1)\bigr)
\right]\notag
\end{align}
becomes 
\begin{align}
\label{eq:heissianmm}
&\frac{c_h^2 \subseq^{-\powerstep+2\powerspa+\powerbatch-1}}{4} \precon\Bigg\{\frac{1}{\batchsizen}
\Bigg[\expect_{\tilde \zeta}[
\frac{\nabla\ell\!\bigl(
\mleglobn+(\spascaln)^{-1}\limglob;\obs{1},\tilde\zeta_1
\bigr)^{\otimes 2}}{\batchsizen}:\bigl(\nabla_{\limglob\limglob} f(\limglob,\newlat_1)-\nabla_{\limglob\limglob} f(\limglob,\limlat_1)\bigr)]
\\
&+
\left(\frac{\batchsizen-1}{\batchsizen}\right)
\frac{1}{\subseq-1}\sum_{i=2}^{\subseq} \expect_{\tilde \zeta}[\nabla\ell\!\bigl(
\mleglobn+(\spascaln)^{-1}\limglob;\obs{i},\tilde\zeta_i
\bigr)^{\otimes 2}]:\expect_{\tilde \zeta}[\bigl(\nabla_{\limglob\limglob} f(\limglob,\newlat_1)-\nabla_{\limglob\limglob} f(\limglob,\limlat_1)\bigr)]
\Bigg] \\
&+
\frac{(\batchsizen-1)}{\batchsizen}
\Bigg[
\frac{1}{{\batchsizen}^{2}} \expect_{\tilde \zeta}[\nabla\ell\!\bigl(
\mleglobn+(\spascaln)^{-1}\limglob;\obs{1},\tilde\zeta_1
\bigr)^{\otimes 2}: \bigl(\nabla_{\limglob\limglob} f(\limglob,\newlat_1)-\nabla_{\limglob\limglob} f(\limglob,\limlat_1)\bigr)]
\\
&+
\frac{2(\batchsizen-1)}{{\batchsizen}^2} \expect_{\tilde \zeta}[\nabla\ell\!\bigl(
\mleglobn+(\spascaln)^{-1}\limglob;\obs{1},\tilde\zeta_1
\bigr):\bigl(\nabla_{\limglob\limglob} f(\limglob,\newlat_1)-\nabla_{\limglob\limglob} f(\limglob,\limlat_1)\bigr)]\\
&\otimes\expect_{\tilde \zeta}[\frac{1}{\subseq-1}\sum_{i=2}^{\subseq} \nabla\ell\!\bigl(
\mleglobn+(\spascaln)^{-1}\limglob;\obs{i},\tilde\zeta_i
\bigr)]
\\
&+
\left(\frac{\batchsizen-1}{\batchsizen}\right)^2 \expect_{\tilde \zeta}\left(\frac{1}{\subseq-1}\sum_{i=2}^{\subseq} \nabla\ell\!\bigl(
\mleglobn+(\spascaln)^{-1}\limglob;\obs{i},\tilde\zeta_i
\bigr)\right)^{\otimes 2}:\expect_{\tilde \zeta}[\bigl(\nabla_{\limglob\limglob} f(\limglob,\newlat_1)-\nabla_{\limglob\limglob} f(\limglob,\limlat_1)\bigr)]\Bigg] \Bigg\}\precon'.
\end{align}

Then we want to write the terms with index $2,\cdots,\subseq$ as summation over $1,\cdots,\subseq$ and terms with index $1$.
\begin{align}
&\frac{1}{\subseq-1}\sum_{i=2}^{\subseq} \expect_{\tilde \zeta}[\nabla\ell\!\bigl(
\mleglobn+(\spascaln)^{-1}\limglob;\obs{i},\tilde\zeta_i
\bigr)^{\otimes 2}]\\
&=\frac{\subseq}{\subseq-1} \left(\tilde I_\star + R_I - \subseq^{-1} \expect_{\tilde \zeta}\nabla\ell\!\bigl(
\mleglobn+(\spascaln)^{-1}\limglob;\obs{1},\tilde\zeta_1
\bigr)^{\otimes 2}\right),
\end{align}
\begin{align}
&\expect_{\tilde \zeta}[\frac{1}{\subseq-1}\sum_{i=2}^{\subseq} \nabla\ell\!\bigl(
\mleglobn+(\spascaln)^{-1}\limglob;\obs{i},\tilde\zeta_i
\bigr)]\\
&= \frac{\subseq}{\subseq-1} \left(\subseq^{-\powerspa} (J_*\,\limglob
+ R_J)
-
\subseq^{-1}
\nabla\ell\!\bigl(
\mleglobn+(\spascaln)^{-1}\limglob;\obs{1}
\bigr)\right)
\end{align}
\begin{align}
&\expect_{\tilde \zeta}\left(\frac{1}{\subseq-1}\sum_{i=2}^{\subseq} \nabla\ell\!\bigl(
\mleglobn+(\spascaln)^{-1}\limglob;\obs{i},\tilde\zeta_i
\bigr)\right)^{\otimes 2}\\
& = \frac{1}{(\subseq-1)^2} \Bigg\{\left(\sum_{i=1}^{\subseq}
\nabla\ell\!\bigl(
\mleglobn+(\spascaln)^{-1}\limglob;\obs{i}
\bigr)\right)^{\otimes2} +\sum_{i=1}^{\subseq} \expect_{\tilde \zeta}[\nabla\ell\!\bigl(
\mleglobn+(\spascaln)^{-1}\limglob;\obs{i},\tilde\zeta_i
\bigr)^{\otimes 2}]\\
&-\sum_{i=1}^{\subseq} \nabla\ell\!\bigl(
\mleglobn+(\spascaln)^{-1}\limglob;\obs{i}
\bigr)^{\otimes 2} -2 \nabla\ell\!\bigl(
\mleglobn+(\spascaln)^{-1}\limglob;\obs{1}\bigr)\left(\sum_{i=1}^{\subseq}
\nabla\ell\!\bigl(
\mleglobn+(\spascaln)^{-1}\limglob;\obs{i}
\bigr)\right)\\
&+2 \nabla\ell\!\bigl(
\mleglobn+(\spascaln)^{-1}\limglob;\obs{1}\bigr)^{\otimes2} - \expect_{\tilde \zeta}[\nabla\ell\!\bigl(
\mleglobn+(\spascaln)^{-1}\limglob;\obs{1},\tilde\zeta_1
\bigr)^{\otimes 2}]
\Bigg\}
\end{align}

So only consider the dominant terms,
\begin{align}
\eqref{eq:heissianmm} \lesssim &\subseq^{-1} \expect_{\tilde \zeta}[
\nabla\ell\!\bigl(
\mleglobn+(\spascaln)^{-1}\limglob;\obs{1},\tilde\zeta_1
\bigr)^{\otimes 2}:\bigl(\nabla_{\limglob\limglob} f(\limglob,\newlat_1)-\nabla_{\limglob\limglob} f(\limglob,\limlat_1)\bigr)] +\subseq^{\powerbatch-1}\|\nabla^{\otimes2}\|_{\infty}\tilde I_\star\\
&+\subseq^{\powerbatch-1-\powerspa}\expect_{\tilde \zeta}[
\nabla\ell\!\bigl(
\mleglobn+(\spascaln)^{-1}\limglob;\obs{1},\tilde\zeta_1
\bigr)\otimes J_\star\limglob:\bigl(\nabla_{\limglob\limglob} f(\limglob,\newlat_1)-\nabla_{\limglob\limglob} f(\limglob,\limlat_1)\bigr)] \\
&+ \|\nabla^{\otimes2}\|_{\infty}\left(\subseq^{\powerbatch-\powerstep-1}(J_*\limglob)^{\otimes2}+\subseq^{2\powerbatch-2}(\tilde I_\star -I_\star) + \subseq^{2\powerbatch-2-\powerspa}J_\star\limglob\otimes \nabla\ell\!\bigl(
\mleglobn+(\spascaln)^{-1}\limglob;\obs{1}\bigr)\right)\\
&\lesssim \|\nabla^{\otimes2}\|_{\infty} \Bigg\{
\subseq^{-1}\expect_{\tilde \zeta} \|\nabla\ell\!\bigl(
\mleglobn+(\spascaln)^{-1}\limglob;\obs{1},\tilde\zeta_1
\bigr)\|^2_{\infty} + \subseq^{\powerbatch-1}\tilde I_\star + \subseq^{\powerbatch-1-\powerspa} \expect_{\tilde \zeta} \|\nabla\ell\!\bigl(
\mleglobn+(\spascaln)^{-1}\limglob;\obs{1},\tilde\zeta_1
\bigr)\|\\
&+\subseq^{\powerbatch-\powerstep-1}(J_*\limglob)^{\otimes2}+\subseq^{2\powerbatch-2}(\tilde I_\star -I_\star) + \subseq^{2\powerbatch-2-\powerspa}J_\star\limglob\otimes \nabla\ell\!\bigl(
\mleglobn+(\spascaln)^{-1}\limglob;\obs{1}\bigr)
\Bigg\}
\end{align}
Since $\powerspa-1 < 0$ and $\powerbatch-1 < 0$, and for $m$ large enough, $\max_{1\le i\le \subseq}
\|\nabla \ell(\globpar^*;\obs{i},\cdot)\|
\le \subseq^{1/p_2}$, $R_5$ vanishes uniformly on $K_1$.

\subsection{$R_6$ (third-order remainder term)}

By the triangle inequality,
\begin{align}
&\subseq^{\powertime}\,
\mathbb{E}\!\left[
\frac{1}{6}\,
\nabla_{\limglob}^{\otimes 3}f(\limglob+S\globincre,\newlat_1)
(\globincre,\globincre,\globincre)
\right] \\
&\le
\frac{27}{6}\,\subseq^{\powertime}\,
\|\nabla_{\limglob}^{\otimes 3} f\|_{\infty}\,
\Bigl(
\mathbb{E}\|\globincrenoise\|^3
+
\mathbb{E}\|\globincreprior\|^3
+
\mathbb{E}\|\globincrell\|^3
\Bigr).
\end{align}

Moreover,
\[
\mathbb{E}\|\globincrenoise\|^3
\le
\subseq^{-\frac{3}{2}(\powerstep+\powerit-2\powerspa)}\,
\left(
\frac{c_h}{2c_\beta}\,
\|\precon\|^{3/2}\,
2^{3/2}\,
\frac{\Gamma\!\left(\frac{d+3}{2}\right)}{\Gamma\!\left(\frac{d}{2}\right)}
\right),
\qquad
\powertime-\frac{3}{2}(\powerstep+\powerit-2\powerspa)<0.
\]
Also,
\[
\mathbb{E}\|\globincreprior\|^3
\le
\left(
\frac{c_h}{2}\,\subseq^{-\powerstep+\powerspa-1}\,\|\precon\|
\right)^3
\left(
\|\nabla\log\prior(\globpar^*)\|
+
L_0\|\mleglobn-\globpar^*\|
+
\frac{L_0(2R_0+2c_0)}{\subseq^{\powerspa}}
\right)^3,
\qquad
\powertime-3\powerstep+3\powerspa-3<0.
\]
Finally,
\[
\mathbb{E}\|\globincrell\|^3
\le
\left(\frac{c_h\|\precon\|}{2}\right)^3
\left(
\subseq^{1/p_2-\powerstep+\powerspa}
+
\subseq^{1/p_3-\powerstep+\powerspa}\,\Phi^{(\subseq)}
+
\subseq^{1/p_3-\powerstep}
\right)^3.
\]
Therefore, $R_6$ vanishes uniformly.
\section{Experimental Details}

\label{app:exp}
This appendix provides additional details on the experimental setups used in
Section~\ref{sec:experiments}, including the synthetic data-generating
distributions and real-data description.
Code to reproduce all experiments is available at \url{https://github.com/shawngyn-stack/LVM_scaling_limits}.

\subsection{Gaussian Mixture Model (GMM) Experiments}
\label{app:gmm_details}

We consider a Gaussian mixture model with $K$ components in $\mathbb{R}^d$ and dataset size $N$.
For each observation $i\in\{1,\dots,N\}$, a latent cluster label is drawn as
\[
z_i \sim \mathrm{Categorical}(\pi), \qquad z_i\in\{1,\dots,K\},
\]
and the observation is generated according to
\[
X_i \mid z_i=k \sim \mathcal{N}\!\bigl(\mu_k,\ \mathrm{diag}(\sigma_k^2)\bigr),
\]
where $\mu_k\in\mathbb{R}^d$ denotes the component mean and
$\sigma_k\in\mathbb{R}_+^d$ parameterizes a diagonal covariance matrix.
The global parameters are $\{\mu_k,\sigma_k\}_{k=1}^K$, while the latent variables are the
cluster assignments $\{z_i\}_{i=1}^N$.
\subsection{Latent Dirichlet Allocation (LDA) }
\label{app:lda_details}

We consider the standard LDA generative model with $K$ topics and a vocabulary of size $V$.
Let $D$ denote the number of documents. For each topic $k\in\{1,\dots,K\}$, we draw a
topic-word distribution
\[
\beta_k \sim \mathrm{Dirichlet}(\beta\,\mathbf{1}_V),
\qquad
\beta_k \in \Delta^{V-1},
\]
independently across $k$. Here $\beta>0$ is a symmetric Dirichlet hyperparameter and
$\mathbf{1}_V$ denotes the all-ones vector in $\mathbb{R}^V$.

For each document $d\in\{1,\dots,D\}$, we draw a document-topic proportion vector
\[
\theta_d \sim \mathrm{Dirichlet}(\alpha\,\mathbf{1}_K),
\qquad
\theta_d \in \Delta^{K-1},
\]
independently across documents, where $\alpha>0$ is a symmetric concentration parameter.
We then sample the document length $L_d$ independently from a discrete uniform distribution
over a fixed range.
Conditional on $\theta_d$, each token $n\in\{1,\dots,L_d\}$ is generated by first sampling
a topic assignment
\[
z_{dn} \mid \theta_d \sim \mathrm{Categorical}(\theta_d),
\]
and then sampling the observed word
\[
w_{dn} \mid z_{dn}=k \sim \mathrm{Categorical}(\beta_k).
\]
We collect each document as a sequence of word indices $w_{d,1:L_d}\in\{1,\dots,V\}^{L_d}$.

In our experiments, the global parameters correspond to the topic-word distributions
$\{\beta_k\}_{k=1}^K$ (equivalently, the matrix $\beta\in\mathbb{R}^{K\times V}$ with rows on the simplex),
while the latent variables are the token-level topic assignments $\{z_{dn}\}$. In the experiments, the document-topic proportions $\{\theta_d\}$ are integrated out,
and Gibbs updates are performed over the token-level assignments $\{z_{dn}\}$. For further details, we refer the reader to \citet{LDA}.

\subsection{Real-World Datasets}
\paragraph{Flow cytometry (GMM).}
We use a real flow cytometry dataset in which each observation corresponds to a single cell.
As input features, we retain only the four fluorescence intensity channels
FL1.H--FL4.H, which measure marker-specific protein expression levels.
All features are standardized prior to model fitting.
Ground-truth cell population labels are available and are used for evaluation.

\paragraph{20 Newsgroups (LDA).}
For topic-modeling experiments, we use the 20 Newsgroups dataset as provided by
\texttt{scikit-learn} \citep{scikit-learn}.
Documents are represented as bags of words after standard preprocessing.

\subsection{Setup of SVI and computational cost}
\new{
We used the default schedule recommended by Hoffman et al. (2013) and verified it converges; results are robust to moderate changes in $\tau_0$ and $\kappa$. The hyperparameters’ values are as follows.}

\new{
GMM: mean-field Normal--Gamma variational family initialized via k-means; minibatch size $256$; prior parameters $\alpha_0=1.0$, $\beta_0=1.0$, $a_0=2.0$, $b_0=2.0$; Robbins--Monro step-size schedule $\rho_t = (\tau_0 + t)^{-\kappa}$ with $\tau_0=10$ and $\kappa=0.7$.}

\new{
LDA (synthetic and real data): semi-collapsed SVI with variational family $q(\pi)=\prod_{k}\mathrm{Dirichlet}(\lambda_k)$ and categorical local factors; minibatch size $64$; $10$ local coordinate-ascent updates per document; symmetric priors $\alpha=\beta=0.1$; same Robbins--Monro schedule.}

\new{
In our synthetic GMM experiments, SVI typically stabilizes within roughly 20 iterations, whereas SGLD--Gibbs requires an initial burn-in period of about 200 iterations, followed by additional iterations used for posterior sampling. On a per-iteration basis, the cost of SGLD--Gibbs is approximately \(0.3\)--\(0.6\times\) that of SVI when \(S=1\), and approximately \(1.6\)--\(3.0\times\) when \(S=5\). Consequently, the overall cost of SGLD--Gibbs depends on the number of retained posterior samples. In our experiments, the total cost needed to obtain stable SGLD--Gibbs estimates was roughly \(5\)--\(10\times\) that of SVI. This additional cost was associated with accuracy improvements of about \(20\)--\(50\%\), together with better calibrated uncertainty quantification.
}

\end{document}